\documentclass{egpubl}
\usepackage{eg2024}

\ConferenceSubmission

\usepackage[T1]{fontenc}
\usepackage{dfadobe}  
\usepackage{cite}  %
\BibtexOrBiblatex
\electronicVersion
\PrintedOrElectronic
\ifpdf \usepackage[pdftex]{graphicx} \pdfcompresslevel=9
\else \usepackage[dvips]{graphicx} \fi

\usepackage{egweblnk} 
\usepackage{amsfonts}
\usepackage{amsmath}
\usepackage{overpic}
\usepackage{booktabs}
\usepackage[shortlabels]{enumitem}
\usepackage{color,soul}
\usepackage{tikz,pgfplots}

\usepackage{cuted}
\usepackage{algorithm2e}
\usepackage{multirow}

\title{Neural Semantic Surface Maps}

\author[L. Morreale \& N. Aigerman, \& V. G. Kim \& N. J. Mitra]
{\parbox{\textwidth}{\centering Luca Morreale$^{1}$\orcid{0000-0001-8933-5082}
        and Noam Aigerman$^{2,3}$\orcid{0000-0002-9116-4662} 
        and Vladimir G. Kim$^{3}$\orcid{0000-0002-3996-6588} 
        and Niloy J. Mitra$^{1,3}$\orcid{0000-0002-2597-0914} 
}
        \\
{\parbox{\textwidth}{\centering $^1$Unversity College London\\
         $^2$University of Montreal\\
         $^3$Adobe Research\\
       }
}
}

\def\eg{\emph{e.g.}} 

\def\ie{\emph{i.e.}}

\def\etal{\emph{et al.}}

\newcommand{\inmap}{\phi}
\newcommand{\outmap}{\Psi}
\newcommand{\source}{\mathbf{A}}
\newcommand{\target}{\mathbf{B}}
\newcommand{\smap}{f_\source}   
\newcommand{\tmap}{f_\target}

\newcommand{\Jloss}{\mathcal{L}_{\text{J}}}

\newcommand{\Coneloss}{\mathcal{L}_{\text{Cones}}}
\newcommand{\Seamloss}{\mathcal{L}_{\text{Seamless}}}
\newcommand{\Smoothloss}{\mathcal{L}_{\text{Smooth}}}

\newcommand{\DA}{\downarrow}

\newcommand{\coloredtext}[2]{\textcolor{#1}{\texttt{\textbf{#2}}}}
\definecolor{LimitationF2H}{rgb}{0.36, 0.64, 1}
\definecolor{LimitationT2K}{rgb}{0.88, 0.17, 0.77}
\definecolor{Dense}{rgb}{1, 0.6, 0.0}
\definecolor{Sparse}{rgb}{0.42, 0.66, 0.31}

\def\poincare/{Poincar\'e}

\def\eg{\textit{e.g.}}
\def\etal{\textit{et al.}~}

\def\ie{\textit{i.e.}}

\newcommand{\abs}[1]{\left\vert#1\right\vert}

\newcommand{\set}[1]{\left\{#1\right\}}
\newcommand{\parr}[1]{\left (#1\right )}

\newcommand{\Real}{\mathbb R}

\newcommand{\eps}{\varepsilon}

\def\mobius/{M{\"o}bius}

\DeclareMathOperator*{\argmax}{argmax}

\newcommand{\change}[1]{#1}
\newcommand{\changeb}[1]{#1}
\newcommand{\changec}[1]{#1}

\begin{document}



\teaser{
\vspace{-0.45in}
\includegraphics[width=\textwidth]{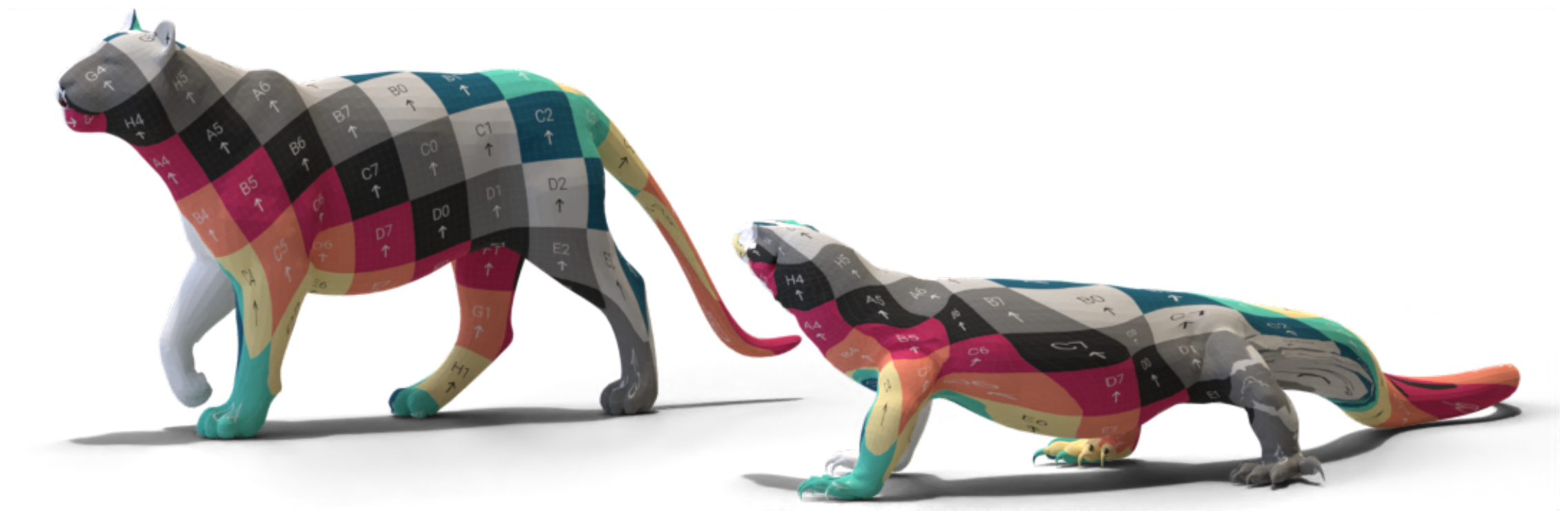}
    \centering
    \caption{   
        Here, we show the extracted map between two non-isometric shapes, \texttt{tiger} and \texttt{iguana}. Although the shapes are highly non-isometric, \eg, lengths of the tail and legs, our method successfully associated these regions between shapes, thus yielding a semantically correct map. This map is seamless by construction, and optimized with no supervision thanks to pre-trained ViT models which can identify semantically corresponding points across shape renderings.
    }
    \label{fig:teaser}
}

\maketitle

\begin{abstract}
We present an automated technique for computing a map between two genus-zero shapes, which matches semantically corresponding regions to one another. Lack of annotated data prohibits direct inference of 3D semantic priors; instead, current state-of-the-art methods predominantly optimize geometric properties or require varying amounts of manual annotation. To overcome the lack of annotated training data, we distill semantic matches from pre-trained vision models: our method renders the pair of untextured 3D shapes from multiple viewpoints; the resulting renders are then fed into an off-the-shelf image-matching strategy that leverages a pre-trained visual model to produce feature points. This yields semantic correspondences, which are projected back to the 3D shapes, producing a raw matching that is inaccurate and inconsistent across different viewpoints. These correspondences are refined and distilled into an inter-surface map by a dedicated optimization scheme, which promotes bijectivity and continuity of the output map.
We illustrate that our approach can generate semantic surface-to-surface maps, eliminating manual annotations or any 3D training data requirement. Furthermore, it proves effective in scenarios with high semantic complexity, where objects are non-isometrically related, as well as in situations where they are nearly isometric.

\end{abstract}

\section{Introduction}

In this work, we propose an automatic method to compute a continuous correspondence between two genus-zero surfaces, represented as meshes. Our core contribution is an approach for computing a \emph{semantic} map that matches semantically corresponding points to one another (e.g., nose to nose, arm to arm, etc.). 

Computing correspondences between domains is a fundamental and highly-researched problem, spanning a wide array of domains such as text snippets~\cite{Hu2019ASO}, audio~\cite{audioSurvey:20}, images~\cite{imageSurvey:21}, or general graphs~\cite{graphMatching:20}. 
In the context of 3D surfaces, establishing such correspondences enables texture or deformation transfer~\cite{defTransfer:04,defTransfer:09}, shape analysis~\cite{eigenFaces:91,humanSpace:16,bogo2014faust,DBLP:journals/corr/PishchulinWHTS15}, and shape space exploration~\cite{HWAG2009SDUMA,glass2022sanjeev,yypm_constrmesh_siga_11}.

Continuous surfaces (2-manifolds), encoded as triangular meshes, remain the most natural and common representation of 3D shapes in graphics and discrete differential geometry. Correspondence between two such surfaces is typically required to be a map that is continuous, one-to-one, onto, and with a continuous inverse, \ie, a \emph{homeomorphism}. Decades of research (see surveys \cite{sahilliouglu2020recent,DBLP:journals/corr/abs-1909-01815}) have been dedicated to tackle the task of mapping between surface pairs. These previous works, being geometric, cannot extract (semantic) maps over the space of homeomorphisms; instead, they have focused on surrogate optimization tasks that minimize some geometric notion of ``distortion'' of the map, \eg, preserving geodesic distances as best as possible. Such distortion-minimizing geometrically-guided maps are, of course, not necessarily semantically meaningful. Thus, a human-in-the-loop approach is usually taken to manually indicate landmark correspondences, which are then used to optimize a map.

Computing semantic homeomorphism faces two main challenges. First, the lack of annotated 3D data inhibits learning high-level semantic priors. 
In contrast, considering the image domain, recent works~\cite{simeoni2021localizing,hamilton2022unsupervised,vaze2022generalized,amir2021deep,oquab2023dinov2} demonstrate that the features of a pre-trained vision transformer~(ViT) are often semantically meaningful and can be used reliably across multiple vision tasks, even on out-of-training image data in a zero-shot setting. 
Second, most 3D representations either hinder or - completely - prevent the computation of bijective inter-surface maps from semantic priors. 
We aim to bridge the semantic matching capabilities of the image domain with the computation of inter-surface maps from potentially noisy correspondences, encouraging continuity and bijectivity. Our core observation is that suitable renderings of the surfaces, without access to surface texture, are already sufficient for image transformers (i.e., ViT) to produce 2D matches that can subsequently be used as fuzzy (i.e., partial and non-injective) maps between the surfaces. Then, we formulate an optimization to aggregate multiple such fuzzy matches obtained from multiview renderings to produce a surface map that best conforms to these fuzzy matches, thereby distilling their semantic priors, see Figure~\ref{fig:teaser}.

\begin{figure}[t!]
    \centering
    \includegraphics[width=\columnwidth]{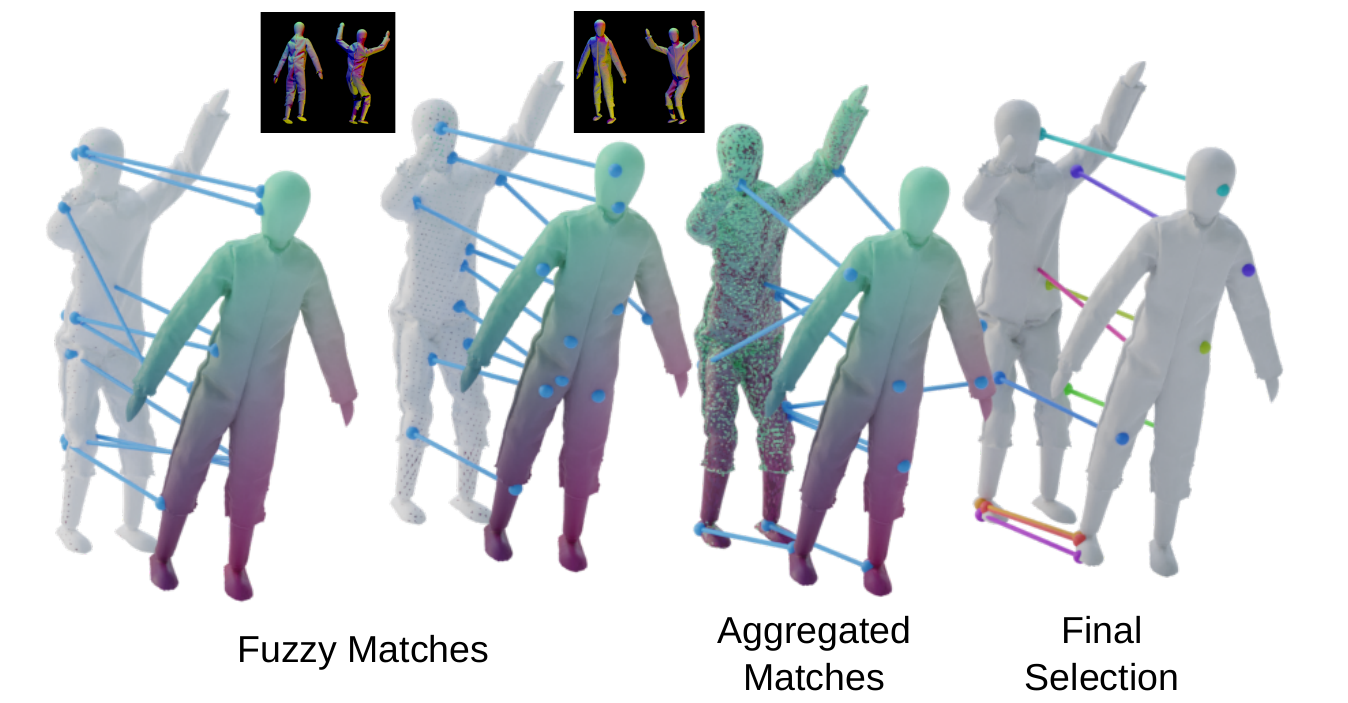}
    \caption{
    \textbf{Fuzzy semantic correspondences.}
    (Left)~We lift 2D image-based correspondences, obtained using a pre-trained vision-transformer~\cite{oquab2023dinov2} on rendered image source/target pairs from sampled views, to obtain fuzzy and spurious 3D (semantic) correspondences. 
    We collect correspondence, shown with coloring and a random set highlighted with lines, from each of the sampled views and aggregate them across views to get aggregated fuzzy matches~(middle), which contain erroneous matching, e.g., thigh getting mapped to the arm. (Right)~We propose an optimization to distill these fuzzy matches into an inter-surface map\change{, here depicting a subset of matches closer than a given threshold ($d<0.1$) wrt the optimized map}.
    }
    \label{fig:matches_vit}
\vspace*{-.2in}
\end{figure}

Specifically, given the fuzzy matches, we utilize Neural Surface Maps (NSM)~\cite{morreale2021neural} to optimize a map between two surfaces. The original NSM framework encodes surfaces using dedicated neural functions, offering a differentiable backbone and avoiding complexities arising from topological changes (i.e., mesh connectivity for different triangulations). However, it has two limitations: it expects the individual surfaces to be cut into disc topologies with the two respective boundaries \textit{already} in correspondence and requires a set of exact landmark correspondences. 
We address the first problem by proposing a seamless Neural Surface Maps~(sNSM) framework, which relaxes the requirement from exact boundary correspondences to only cone-point matchings.
We address the second problem by optimizing a custom objective that encourages the image of a specific point to best accommodate the fuzzy (semantic) matches while identifying and disregarding outliers (see Figure~\ref{fig:matches_vit}). The resultant optimization problem is solved using gradient descent, simply through PyTorch's SGD optimizer.

Through quantitative and qualitative experiments, we evaluate our ability to match upright object pairs with varying levels of isometry for objects from the same semantic class and across different ones. We also compare ours to competing surface map extraction algorithms. In summary, our main contributions are:
\begin{itemize}
    \item proposing a fully automatic algorithm for extracting semantic maps between upright shape pairs; 
    \item sampling and integrating a set of image-based
    correspondences to form fuzzy object space correspondence maps;
    \item extending Neural Surface Maps framework to seamless maps that can work with fuzzy, and potentially noisy guidance, to distill semantic maps; and
    \item demonstrating, via extensive evaluation and comparisons, that the algorithm yields semantically valid maps for both isometrically and non-isometrically related shape pairs. 
\end{itemize}

\section{Related Works}

\subsection{Shape Matching}
Shape matching and correspondence estimation have been widely studied in geometry processing. In the simplest setting where rigid transforms relate shape pairs, iterative closest points methods~\cite{rusinkiewicz2001efficient,zinsser2003refined} are often used for local refinement while relying on deep learned features to identify good correspondences to provide an initial global alignment~\cite{gojcic2019perfect,deng2018ppfnet,wang2019deep}. Directly optimizing for distance preservation is computationally challenging for surfaces~\cite{Bronstein06a,nonRigid:qixing:2008} discretized as dense triangulations. Optimal transport  has been used to optimize over relatively coarse shape representations~\cite{Solomon16}.

Functional maps compute a fuzzy correspondence by aligning the spectral basis of two shapes that are related by linear transforms for near-isometric deformations~\cite{ovsjanikov2012functional,corman:functional:2017}. The approach offers elegant machinery to compute shape correspondence and has been extended to handle partial matching~\cite{Litany18partialfn}, learn spectral alignment~\cite{litany2017deep,donati2020deep}, find multiple low-distortion maps~\cite{ren2020maptree}, and leverage optimal transport methods~\cite{pai2021fast} (c.f.~course notes~\cite{Ovsjanikov17survey} for other variations). 
While variants have been proposed to `project' functional maps to point-to-point maps, such approaches~\cite{Ezuz17_deblurring} do not explicitly ensure bijectivity of the maps.

Restricting optimizations directly to surfaces poses a challenge since the most common surface representation, a triangle mesh, does not easily lend itself to continuous optimization. Schreiner \etal \shortcite{schreiner2004inter} optimize surface-to-surface maps via direct mesh-mesh intersection, which leads to a combinatorial problem due to the surface discretization into triangles. Subsequent approaches represent maps by parameterizing meshes to a common domain such as a plane~\cite{aigerman2014lifted,schmidt2019distortion} or a sphere~\cite{Asirvatham05,bijective:EG:23}, which can give a bijective final map, but still does not offer a natural way to optimize using inter-surface distortion. Notably, Schmidt \etal \shortcite{bijective:EG:23} propose a coarse-to-fine optimization strategy, defined between spheres, that uses incremental re-meshing to significantly speed up the optimization and improve the quality of the map, yet relies on few manual annotations. %

Others treated inter-surface mapping as a problem of exploring a space of low-distortion maps, for example, blending across conformal maps, Blended Intrinsic Maps~(BIM)~\cite{Kim11BIM} were explored and interpolated. The recently proposed Enigma~\cite{Edelstein20Enigma}  leverages genetic algorithms to achieve state-of-the-art results. In another notable effort, the  MapTree~\cite{ren2020maptree}, authors explore multiple functional maps between near-isometric shape pairs. They propose a fully automatic method that reveals multiple and diverse maps by first enumerating map variations and then optimizing them to extract dense pointwise maps. 

Parallel to this work, Abdelreheem \etal~\cite{abdelreheem2023zero} proposed a pipeline to extract coarse correspondences between shapes through pre-trained networks. In particular, the authors use Blip2~\cite{li2023blip} to extract a shape class description, \eg, human or person; such class is then used to extract semantic descriptions for shape subparts, \eg, leg or arm. Finally, SAM\cite{kirillov2023segment} extracts features based on these keywords and renderings to extract features for each shape. The resulting features are then used for co-segmentation and shape correspondences. The latter is achieved through the functional map framework~\cite{ren2018continuous}, which natively produces a fuzzy map, and thus may not sufficiently reduce, or refine, the fuzziness of the initial set of correspondences.

Instead of functional maps, which assume that a linear spectral map exists and is less suitable for optimizing bijectivity, we build on NSM~\cite{morreale2021neural} that maps shapes through common 2D domains. However, the original technique implicitly assumes shapes to have corresponding boundaries, making it unsuitable for fully automatic mapping. Thus, we modify the formulation to be seamless across any cut boundary, allowing for arbitrary non-matching cuts on different surfaces. We rely only on DinoV2\cite{oquab2023dinov2} to extract fuzzy features from shape renderings to produce point-level (rather than part-level) correspondence priors.

\subsection{Image-based Shape Analysis}

Image-based representations are commonly used for 3D shape analysis tasks, such as classification~\cite{su15mvcnn}, segmentation~\cite{MvDeCor2022,Kundu:2020:multiview_fusion,decatur20223d}, or matching~\cite{Huang:2017:LMVCNN}, where a shape is first rendered from multiple viewpoints, the resulting images are analyzed with 2D neural networks, and then the output is aggregated on the 3D shape via additional optimization~\cite{Kalogerakis:2017:ShapePFCN}. While these methods often start with pre-trained 2D neural networks, they require additional fine-tuning with 3D supervision and thus can only work on categories of shapes with labeled 3D data. Indeed, Genova et al.~\shortcite{Genova:2021:3D_seg_2D_sup}, train a 3D segmentation technique by using a 2D method to produce pseudo labels. In a concurrent effort, \cite{abdelreheem2023satr} describes a training-free approach for 3D shape semantic segmentation using pre-trained visual transformers.
Our approach exploits a similar intuition to extract semantic shape correspondences, distilling an inter-surface map from them, thus \textit{without} any 3D supervision.

\begin{figure*}[t]
    \centering
    \includegraphics[width=\textwidth]{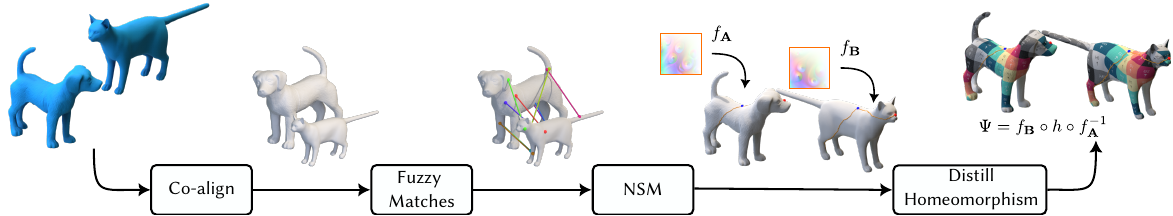}
    \caption{
    \textbf{Overview.}
    Starting from a pair of upright genus-zero surfaces, we automatically distill an inter-surface map from a set of fuzzy matches. First, we align the input shapes, then extract a set of fuzzy matches through DinoV2~\cite{oquab2023dinov2} semantic visual features. We use these features to independently cut the two meshes and then optimize a (seamless) map between them.
    }
    \label{fig:pipeline}
\end{figure*}

\subsection{Visual Features}
The use of pre-trained CNNs features marked a vital milestone for computer vision tasks, such as object detection and segmentation \cite{girshick2014rich}, or image synthesis \cite{gatys2016image,shocher2020semantic}. These network representations encode a wide range of visual information from low-level (statistical) features, (\eg, edges, auto-correlation matrices), 
object parts, and structure~\cite{olah2017feature,carter2019activation,MoGuerreroEtAl:StructureNet:SiggraphAsia:2019}. However, these methods~\cite{luo2016understanding} usually are restricted to local (CNN) neighborhoods or pre-authored nonlocal receptive field, and ignore long-range dependencies \cite{wang2018non}. %

Vision Transformers \cite{dosovitskiy2020image}, dubbed ViT, belong to a family of recent and powerful neural architecture that can discover both local and nonlocal relations. A noteworthy example is DINO-ViT~\cite{caron2021emerging} that trains a transformer network through self-distillation and uses its features in several tasks, \eg, image retrieval and object segmentation. Several works demonstrated the utility of Dino-ViT internal representation as a black box~\cite{simeoni2021localizing,wang2022self} for tasks such as semantic segmentation \cite{hamilton2022unsupervised} and category discovery \cite{vaze2022generalized}. Amir \etal\shortcite{amir2021deep} study  these features and use them to solve vision tasks, such as image correspondences, in zero-shot settings. Recently, Oquab \etal\shortcite{oquab2023dinov2} extended Dino-ViT, introducing Dinov2, showcasing enhanced feature semantic interpretability compared to the original version, and also exhibiting broader applicability.

\section{Method} \label{sec:method}

We now detail our framework (see Figure~\ref{fig:pipeline}) for an automatic inter-surface map. We assume to be given two upright 3D surfaces, $\source$ and $\target$, in arbitrary relative poses. The majority of meshes from online repositories, such as the 3D Warehouse, TurboSquid, or Sketchfab, satisfy this requirement, alternatively, methods like \cite{pang2022upright,poursaeed2020self} can be used as pre-processing. 
We assume both shapes to have zero genus, although the method can be extended to higher genus surfaces.
We aim to compute an inter-surface map $\outmap:\source\leftrightarrow\target$ guided by visual semantics. Our framework proceeds in three stages: 
\begin{enumerate}[(i)]
    \item Given two shapes, $\source$ and $\target$, that are assumed to be oriented upright, we automatically align them using semantic matches.
    
    \item We aggregate fuzzy matches (i.e., general matching of pairs of points which is neither 1-to-1, onto, nor maps all points on the source surface) between the surfaces by applying 2D matching techniques to renderings, over multiple views. 
    
    \item Optimize a surface map that best agrees with the fuzzy semantic matches while handling outliers. 
\end{enumerate}

\subsection{Semantic Shape Alignment}
\begin{figure}[!ht]
    \centering
    \includegraphics[width=\columnwidth]{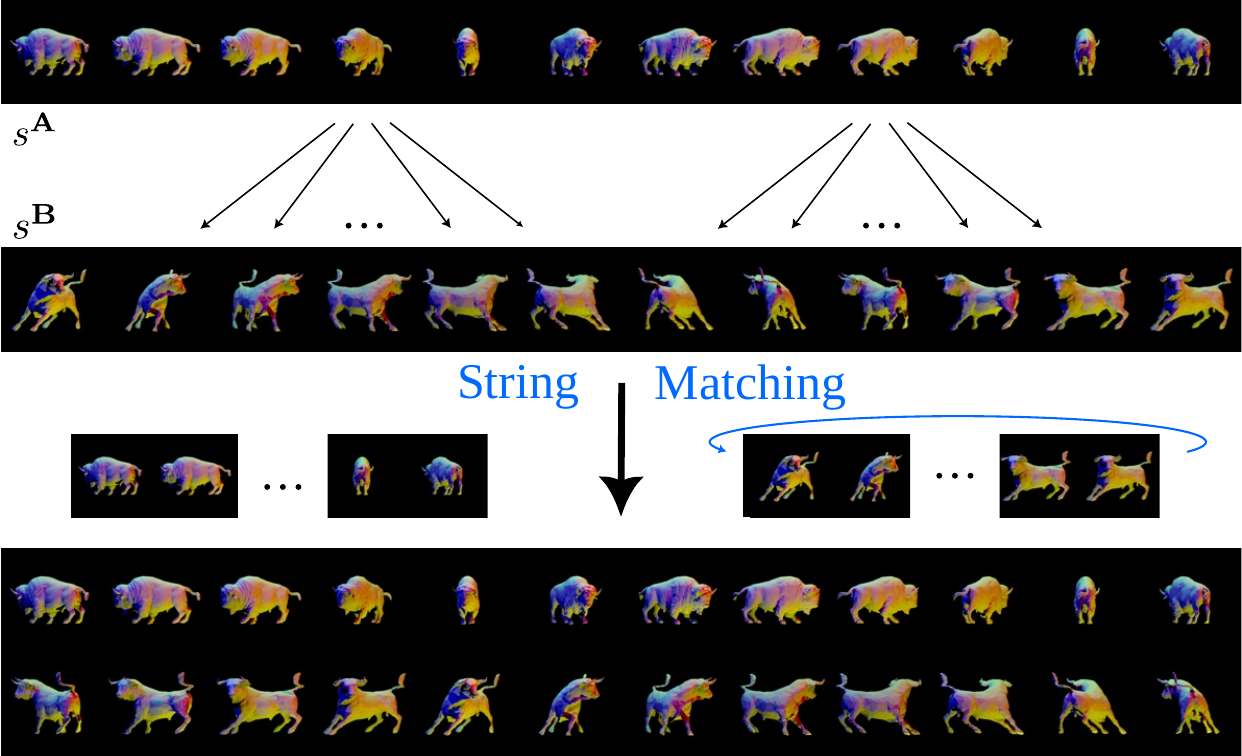}
    \caption{
    \textbf{Co-aligning input surfaces:} \label{fig:alignment}
    Starting from a pair of upright meshes ({\texttt{bison}} and {\texttt{bull}} in this example), we render $12$ views around them ($s^\source$ and $s^\target$). Then, we extract DinoV2 features from each rendering independently and match these features as a string-matching problem. Specifically, we optimize over a cyclic shift of the rendered views (i.e., one degree of freedom) to maximize agreement of image-based semantic correspondences.
    }
    \vspace{-0.15in}
    
\end{figure}

Given two upright shapes, $\source$ and $\target$, we first align them to have the same orientations. We achieve this by casting this problem as (semantic) circular string matching between shape renderings: given two `strings' -- sets of renderings -- of the same length, we find the global rotation $\mathbf{R}$, about the upaxis, to best align one string with the other. Intuitively, we order one sequence to convey semantic information in the same order as the other, see Figure \ref{fig:alignment} for an overview.

First, we render each mesh from 12 viewpoints around it, $R_i^\source$ and $R_i^\target$ (see Subsection \ref{subsec:rendering} for a discussion on rendering). These images constitute the two strings $s^\source := \{ R_i^\source \}$ and $s^\target := \{ R_i^\target \}$.
Then, we extract a set of   DinoV2~\cite{oquab2023dinov2} features for each image.
Finally, we compute the alignment score for the 12 possible rotations as the total number of "Best Buddy matches"~\cite{dekel2015best} between the two strings of features. We pick the (relative) rotation with the highest score as the rotation and use it to co-align the shapes.

\subsection{Distilling Fuzzy 3D \changeb{Correspondences} via Visual Semantics}
Next, we extract fuzzy matches from renderings, taken from different viewpoints, of the aligned surfaces. Each such viewpoint $V$ results in a pair of rendering that we use to define a fuzzy match $\inmap^V:=\left(p^V_i,q^V_i\right)_{i=1}^n$ with $p\in\source,q\in\target$, which consists of pairs of corresponding points on $\source$ and $\target$. 

Although matches are imprecise or inaccurate, we assume that these imprecisions balance out, leading to approximately correct matches. Embracing this assumption, we leverage it as a guiding principle during map optimization.

\paragraph*{Computing rendering matches.} 
Given a viewpoint $V$, we render the two untextured surfaces from that viewpoint to get two renderings, $R^\source_V$ and $R^\target_V$. To extract matches, we take inspiration from recent methods that leverage deep image features from \cite{oquab2023dinov2} for matching 2D images and design a method for extracting dense visual matches. 
Specifically for each image patch processed by Dinov2, we extract a feature vector with $\lambda^\source_{i}$ and $\lambda^\target_i$ being the features of rendering of $R^\source_V$ and $R^\target_V$, respectively. Then, we segment foreground/background through PCA and compute the cosine similarity between all pairs of source and target patch foreground features, as score 
\begin{equation}
S_{ij} = \left<\lambda^\source_i,\lambda^\target_j\right>.
\end{equation}
Finally, we define the match of patch $i\in R^\source_V$ as the patch $j\in R^\target_V$ with the highest cosine similarity, and vice versa, the match of patch $j\in R^\target_V$ as the patch $i\in R^\source_V$ with the highest cosine similarity. In summary, the pair $(i,j), i\in R^\source_V,j\in R^\target_V$ is a match, if 
\begin{equation}
S_{ij} = \max_k S_{ik} \text{ or } S_{ij} = \max_l S_{lj}.
\end{equation}
We transform a match from patch level to pixel level, as the patch size is known.
In contrast to ours, \cite{amir2021deep} selects only "Best Buddy" matches \cite{dekel2015best}, augments features with binning, and does not segment foreground/background through PCA features. Although \cite{amir2021deep} produces a more expressive set of features and possibly a more reliable set of fuzzy matches, we found it time-consuming (2hrs in our settings), and our experiments did not provide sufficient justification for such a design choice.

Given dense 2D matches in an image, we lift (unproject) each pixel to the 3D mesh by performing ray intersection between that pixel's corresponding ray from viewpoint $V$ and the 3D mesh $\mathbf{T}$, thereby associating every 2D pixel with a point on the surface, represented as barycentric coordinates at the triangle the ray intersects. The fuzzy \changeb{correspondences} are thus pairs of matching 3D points (represented as barycentric coordinates on triangles): \changeb{$\inmap^V:=\left(p^V_i,q^V_i\right)_{i=1}^n$}.

We repeat this process from multiple viewpoints and obtain a collection $\set{\inmap^i}_{i=1}^k$ of fuzzy \changeb{correspondences}. Our final task is to distill them to produce an automatic map.

\subsection{Aggregating the Fuzzy \changeb{Correspondences} to an Inter-surface map} \label{subsec:aggregate}
Given the fuzzy \changeb{correspondences}, we wish to optimize a continuous map $\outmap$ between $\source$ and $\target$ using a differentiable loss that encourages agreement with the fuzzy \changeb{correspondences}. 

Our final goal is thus to devise an optimization scheme that will lead to a map $\outmap:\source\leftrightarrow\target$ which balances smoothness with the number of respected correspondences. \changec{To achieve this goal, we compare each point's image with its designated corresponding point from $\inmap^i$ with the L1 norm. Then, for a set of corresponding pairs $(p_j,q_j)$, we minimize the average error as follows}:
\change{
\begin{equation}
    \label{eq:real_optimization}
    \mathcal{L}_\text{Matches} = \frac{1}{N} \sum_{j=1}^N \|\outmap(p_j)-q_j\|_{1} ,
\end{equation}
}
where $N$ is the number of correspondences. \changec{By averaging these distances, we encourage sparsity of correspondences.}
To optimize \change{Eq. \ref{eq:real_optimization}}, we adopt a recent method for optimization of the surface map, Neural Surface Maps~(NSM)~\cite{morreale2021neural} as described next.

\paragraph*{Seamless Neural Surface Map.}
We follow NSM's paradigm: we first parameterize each one of the two cut surfaces via SLIM~\cite{rabinovich2017scalable} into a square $D\in\Real^2$ to get two bijective seamless parameterizations, $P_\source:\source\leftrightarrow D, P_\target:\target\leftrightarrow D$. Then, we fit a neural network to each of the two parameterizations' inverse, $\smap\approx P_\source^{-1}$, $\tmap\approx P_\target^{-1}$. Finally, using another neural network that maps the square to itself, $h$, we can define the inter-surface map $\outmap = \tmap\circ h\circ\smap^{-1}$. By optimizing solely the parameters of $h$ while maintaining its bijectivity, and holding the overfitted networks $\smap,\tmap$ fixed, NSM enables optimization over the space of maps between the two surfaces.

\begin{figure}[b!]
\centering
         \includegraphics[width=\columnwidth]{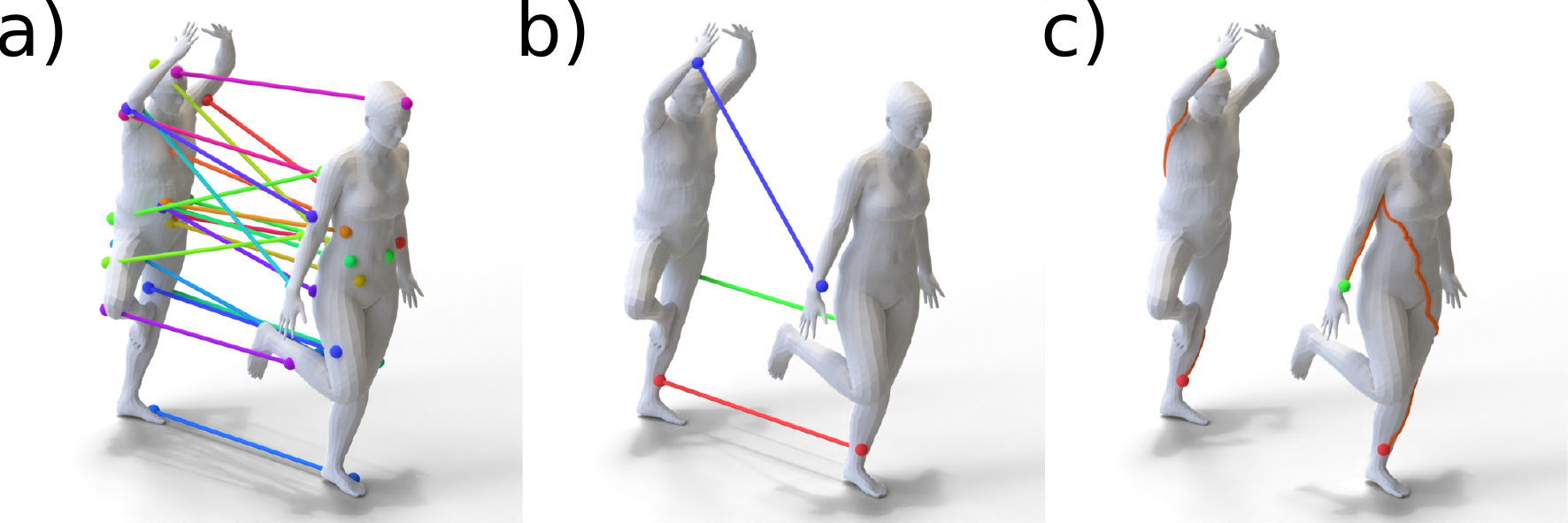}
    \caption{
    \textbf{Cutting through cone points.}\label{fig:autocut}
    We collect a set of spurious and noisy matches~(a). Then, we select the most reliable $K=3$ \changeb{correspondences}~(b). Finally, using these \changeb{correspondences} as cut endpoints, or cones, we cut the two meshes independently~(c). Note how the cut differs in the two shapes: the man is cut through the back, while the woman is cut through the front. Refer to Sec.~\ref{par:cones} for details. 
}
\end{figure}

\begin{figure}[h!]
\centering
    \includegraphics[width=\columnwidth]{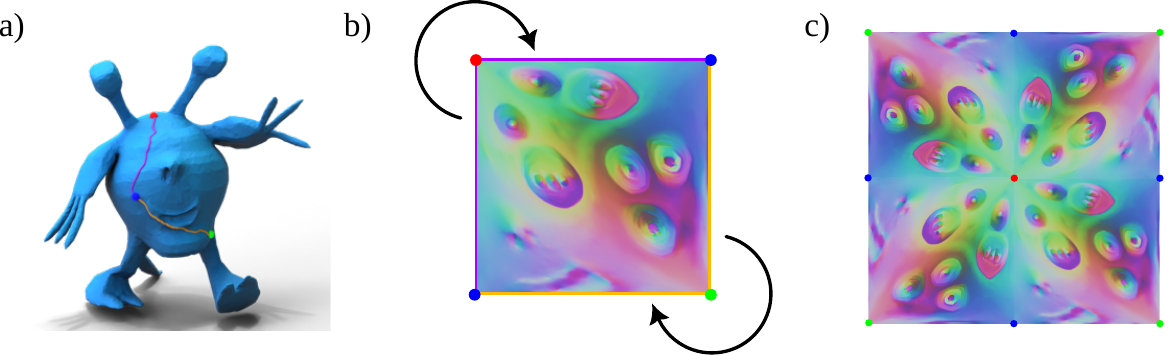}
    \caption{
    \textbf{Seamless cuts.}\label{fig:seamless}
    To parametrize a genus-zero mesh~(a) we cut and map it to a disc topology, cut visualized as in~(b). The two corresponding sides of the cut match perfectly, i.e., when we connect the two parts, the map remains continuous across the cut~(c).
}
\end{figure}

As we cannot guarantee corresponding cuts between genus 0 meshes, see cut examples in Figure \ref{fig:autocut}, we relax the boundary-matching constraint in the original NSM and extend it to support seamless maps. Intuitively, a borderless, or seamless, parameterization is a 2D-3D mapping that is independent of the choice of cut path, given a set of $K$ boundary points. In other words, the map emerging from the parametrization has several equivalent maps with different boundaries, see Figure \ref{fig:seamless}(c). Only the $K$ points, referred to as cones, remain constant and must have the same mapping across all equivalent maps.
Mathematically, a seamless parametrization is a mapping equipped with homotopic cuts (i.e., the cuts can be changed homotopically but the produced mapping will stay the same). In particular, for three cones on a sphere, all cuts are homotopic, and thus the embedding is independent of the cut choice. Please refer to \cite{aigerman2015seamless} for more details.

Furthermore, the class of seamless parametrization requires a specific type of cut such that triangles, or points for that matter, can be mapped to the other side of the cut by a family of transformations $\mathcal{R}$.
In terms of NSM, a seamless map requires matching corresponding cones while the boundary is allowed to move. Thus, the accuracy required to define the cut path, and hence the 2D boundary, through fuzzy matches is reduced, \eg,~see Figure \ref{fig:autocut}(c) for cut paths. Below, we first detail how we extract corresponding cones, then describe a seamless map.

\paragraph*{Cones.} \label{par:cones}
To identify cones, we first aggregate the fuzzy \changeb{correspondences} by counting for each triangle $F^\source_i\in\source$, how many fuzzy \changeb{correspondences} associate it with triangle $F^\target_j\in\target$, yielding a large sparse matrix $M$ such that its $(i,j)$ entry is the total count for \changeb{correspondences} of $F^\source_i$ to $F^\target_j$,
\begin{equation}
    M_{ij} = \sum_k \left|\left\{(p,q) \text{ s.t. }p\in F^\source_i,q\in F^\target_j,(p,q)\in\inmap^k\right\}\right|,
\end{equation}
where $|\cdot |$ stands for the cardinality of the set. 
Next, we consider $M$ as the adjacency matrix of an edge-weighted graph, with $M_{ij}$ being the weight on edge $(i,j)$. Then, through bipartite graph matching~\cite{hopcroft1973n} we obtain a matching, \ie, a list of pairs $(i_k,j_k)$, s.t.~$ i,j = \argmax\limits_{i,j} M_{i, j}$. We select the $K=3$ \changeb{correspondences} $(i,j)$ with the highest $M_{ij}$ values, such that the geodesic distance - averaged between the two shapes - between all $K$ points is at least $\tau=0.3$. Finally, we use these landmarks as the cut's endpoints and the midpoint.

\paragraph*{Seamlessness.}
Since we cannot rely on the cut quality, we reformulate the neural map $h$, which we optimize to define the map $\outmap$, to support seamlessness. This constrains the map to work on shape pairs with the same genus. Furthermore, the definition of the seamless map changes based on the genus. For a sphere, $h$ changes to $\tilde{h}$:
\change{
\begin{equation} \label{eq:seamless_map}
\resizebox{0.9\hsize}{!}{$
    \tilde{h} = \Bigl\{ x \rightarrow T \cdot h(x) + \eta \, | \, T = \left(  
    \begin{array}{cc} 
    a & -b \\ 
    b & a 
    \end{array} 
    \right) \in \Real^{2 \times 2} , \eta \in \Real^{2 \times 1} \Bigr\}
$}
\end{equation}
}
for all points mapped outside the domain $D$, which rotates around the cone $c_i$ ($\eta$) of a rotation $R$ ($T$). 

To achieve seamlessness $h$ must perfectly match cones $c_i$ to their ground truth $\tilde{c}_i$. Therefore, we formulate such a constraint by penalizing the deviation of mapped cones, $h(c_i)$, to their ground truth position: 
\begin{equation}
    \label{eq:cones}
    \Coneloss = \left\Vert h(c_i) - \tilde{c}_i \right\Vert .
\end{equation}
In the case of spheres, we have 3 cones, of which one is duplicated, thus one for each vertex of the square in the square domain $D$. In the case of torus, a single point is duplicated 4 times, corresponding to all 4 square vertices.

A second condition for seamlessness concerns the duplicated points on the boundary. In the case of spheres, each point on the boundary $p_1$ has a corresponding point $p_2$ which is a rotation of $90^\circ$ with respect to one of the cones. For the case of a sphere, we formulate the constraint as the following energy:
\begin{equation}
    \label{eq:seamless}
    \Seamloss = \left\Vert h(p_1) - R \cdot \left( h(p_2) - c_i \right) + c_i \right\Vert,
\end{equation}
where $c_i$ is the cones wrt $p_2$ undergoes a rotation $R$ to be a clone of $p_1$. Note, the rotation can either be $\pi/2$ or $-\pi/2$. In the case of a torus, $p_2$ is on the opposite side of the boundary of $p_1$, i.e., the transformation being a translation along $x$ or $y$.

\begin{figure*}
    \centering
    \includegraphics[width=\textwidth]{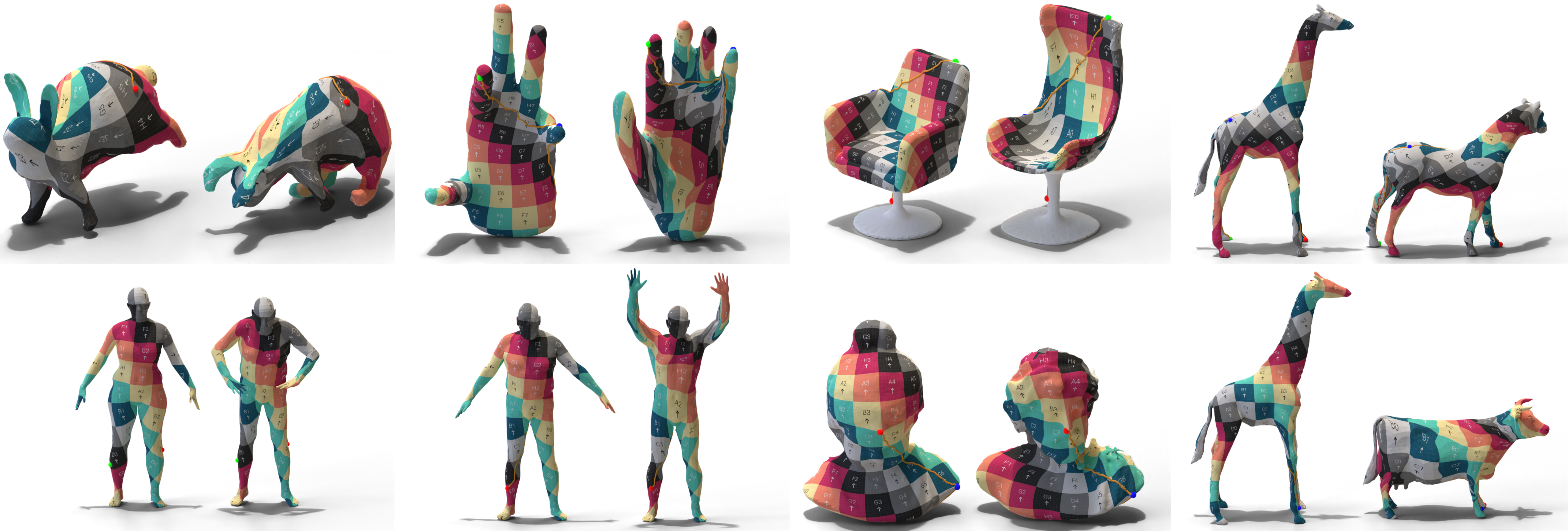}
    \caption{
    \textbf{Results.}
    Automatic maps extracted by the optimization on various surface pairs using aggregated fuzzy correspondences. Colored landmarks and paths show \emph{automatically} selected cones and cuts by our method.
    The rabbit, hands, humans, and heads examples represent near isometric pairs with pose variations; the chairs, giraffe-horse, giraffe-cow examples produce non-isometric mappings with spatially varying distortions. Note the semantic nature of the extracted maps. No explicit energy term was used to encourage the maps to be isometric. 
    }
    \label{fig:resultsPage}
\end{figure*}

\paragraph*{Optimization energies.}
We follow NSM and encourage the map to be bijective through a loss term that prevents the map $h$'s Jacobian $J_p$ at every point $p\in D$ from having a negative determinant:
\begin{equation}
    \label{eq:injectivity}
    \Jloss = \int_D \max \parr{- sign\parr{\abs{J_p}} e^{-\abs{J_p}}, 0}.
\end{equation}
Thus, encouraging\change{, but not guaranteeing,} continuity and bijectivity of the map. %

To cope with the sparsity of fuzzy matches and obtain a well-defined map in undefined regions, we use an energy term that encourages smoothness and prevents large distortion:
\begin{equation}
    \label{eq:symmetric_dirichlet}
    \Smoothloss = \int_D \left\Vert J_p^{\outmap} - J_{p^\eps}^{\outmap} \right\Vert,
\end{equation}
where $J_p^{\outmap}$ is the Jacobian at a point $p$ of the map $\outmap$. While $p^{\eps}$ is the point $p$ perturbed by $\eps \sim \mathcal{N}(0,0.1)$ through barycentric coordinates. Intuitively, we want the Jacobian of the map to change slowly.
Note, in NSM \cite{morreale2021neural}, the authors used Symmetric Dirichlet \cite{rabinovich2017scalable} in a similar context, however, this energy promotes isometric maps rather than smooth ones. Such behavior can actively damage the map optimization and force it to ignore certain - correct - matches, while we aim to attend to unregularized areas.

\paragraph*{Total energy.}
Our total loss is expressed as:
\change{
\begin{equation} \label{eq:loss}
\begin{split}
    \mathcal{L} = \alpha_1 \mathcal{L}_\text{matches} + \alpha_2 \Jloss + \alpha_3 \Coneloss + \\
    \alpha_4 \Seamloss + \alpha_5 \Smoothloss ,
\end{split}
\end{equation}
}
where $\alpha_1 = 10^4 $, $\alpha_2 = 10^6 $, $\alpha_3 = 10^6 $, $\alpha_4 = 10^6 $, and $\alpha_5 = 10^{-1} $ in all experiments. \change{These hyper-parameters were selected experimentally.}
We optimize network weights $h$ using this loss, and to alleviate the impact of incorrect matches; we incrementally drop those that strongly disagree with the current map, \ie, $20\%$ of matches with the highest Euclidean distance. Experimentally, this explicit filtering reduces the impact of incorrect matches, thus preventing the network from getting stuck in incorrect energy minima due to the presence of inaccurate matches.

In summary, the proposed pipeline optimizes for an inter-surface map \emph{automatically} between a pair of upright shapes. First, the input mesh pair is automatically aligned, then through pre-trained ViT~\cite{oquab2023dinov2} we extract a large set of semantic fuzzy matches between them. Finally, we distill an inter-surface map; this step is fundamental to filter out any incorrect matches, enhancing the overall accuracy and reliability of the resultant map. Next, we quantitatively and qualitatively evaluate the quality of the distilled map.

\subsection{Rendering Settings} \label{subsec:rendering}
We render shape pairs and use these images with Dino-ViT2~\cite{oquab2023dinov2}; this model is known to be forgiving in cases of image variation. As the shape alignment is unknown, we render an object-centric scene with a fixed perspective camera and $5$ point lights aimed at the shape. Different points of view are obtained by rotating \emph{only} the shape by fixed increments, while the rest of the scene (\ie, camera and lights) stays fixed. We set up the scene to ensure the entire object is visible by the camera's field of view.

To boost the matching capabilities of Dino-ViT and aid it in distinguishing left from right, top from bottom, while enhancing scene details, we strategically position colored lights around the object in a half-dome fashion. Specifically, we employ five colored point lights (red, blue, green, yellow, and white) for this purpose. As depicted in Figure \ref{fig:alignment}, corresponding regions in the images exhibit similar colors; for instance, the right part of the images tends to appear reddish due to illumination from a red light source. In cases involving textured meshes, we replace the colored lights with white ones.

\changeb{
\subsection{Implementation  details}
Following NSM \cite{morreale2021neural}, we never require to compute $\Psi$ to optimize $\mathcal{L}$.
To evaluate $\mathcal{L}_\text{Matches}$ we first compute the barycentric coordinates of $p_j$ and convert it to a point in the square, $p_j^{2D} \in \Real^2$. Then, this point is mapped forward through $f^\target \circ h$ and used to compute the error as $\|f^\target(h(p^{2D}_j)) - q_j\|_{1}$ for each correspondence. Furthermore, as the number of correspondences is extremely large $N \approx 65k$, at runtime we estimate $\mathcal{L}_\text{Matches}$ on a subset ($M << N$).
We follow a similar strategy for $\mathcal{L}_{Seamless}$ by precomputing a set of $p_j^{2D}$ from the boundary and pushing them forward $f_B \circ h$. Differently, $\mathcal{L}_{Smooth}$ and $\mathcal{L}_{J}$ require only the computation of Jacobians which can be estimated from forward maps. Finally, $\mathcal{L}_{Cones}$ penalizes the prediction of known 2D points, thus requiring only $h$. During the evaluation, we rely on $h$ using $P_B \circ h \circ P_A^{-1}$, see supplementary material for more detail.
}

\changeb{
In all experiments, we define the neural map ($h$) as a 4-layer residual MLP of 128 neurons each, while neural surfaces ($f$) are always 8 layers residual MLP with $256$ neurons.
While training, we sample $1024$ points to enforce injectivity and smoothness, and $128$ points on the boundary to enforce seamlessness and $M=128$ correspondences in each iteration.
}

\begin{figure*}
    \centering

    \includegraphics[width=0.98\textwidth]{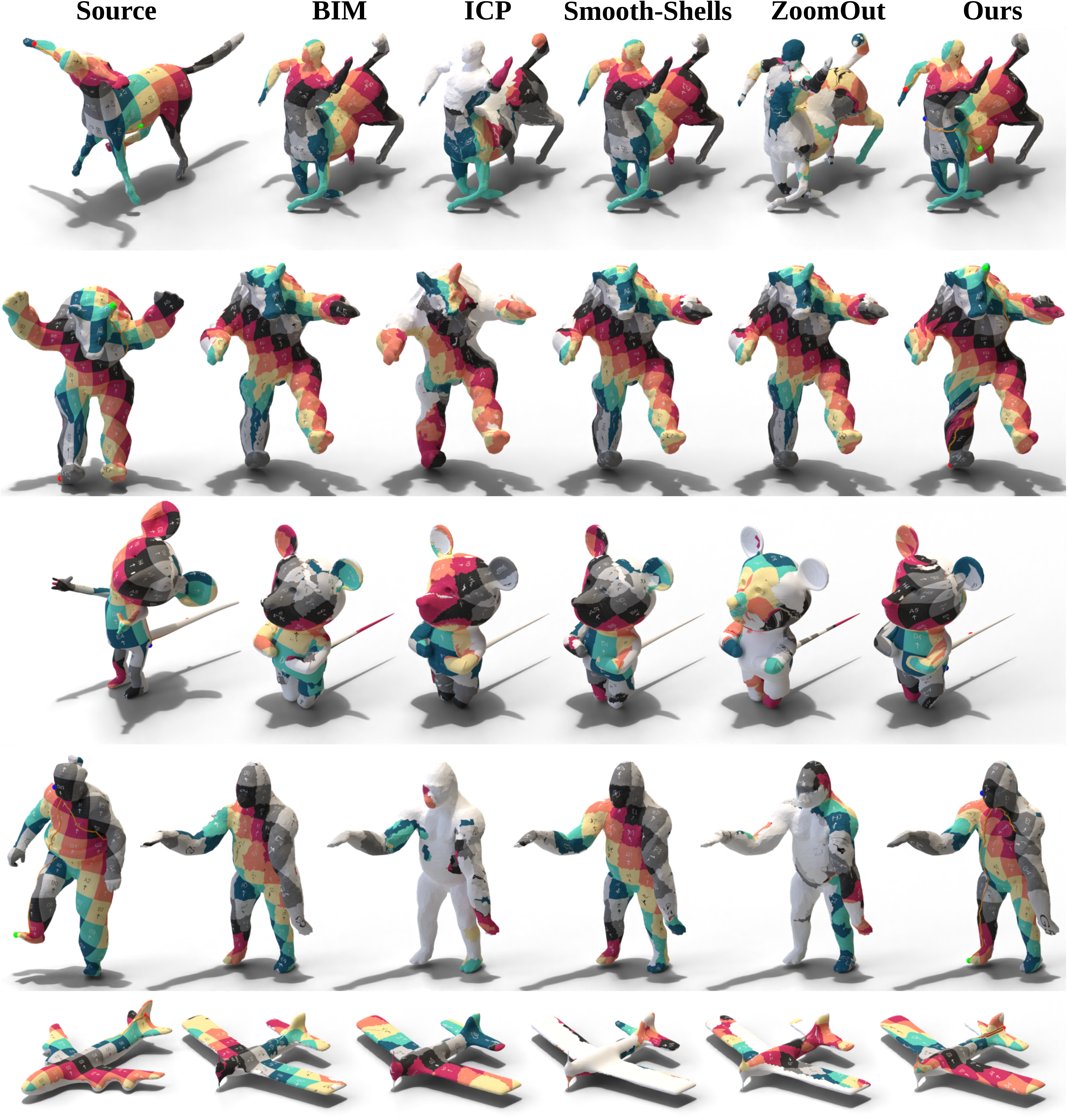}
    \caption{
    \textbf{Comparisons.} \label{fig:comparison}
     Left-to-right: Source model, results using BIM~\shortcite{Kim11BIM}, ICP, Smooth-shells~\shortcite{eisenberger2020smooth}, ZoomOut~\shortcite{melzi2019zoomout}, and Ours. Although geometric methods produce good maps, they often yield discontinuous maps, e.g., see the wings of planes. Ours explicitly encourages continuity and \change{bijectivity}. Colored landmarks and paths show \emph{automatically} selected cones and cuts by our method. 
     Note that our maps are continuous across the cut seams. No explicit energy term is used to encourage isometric maps. 
     }
    
\end{figure*}

\begin{figure*}[t]
\centering 
    \includegraphics[width=\textwidth]{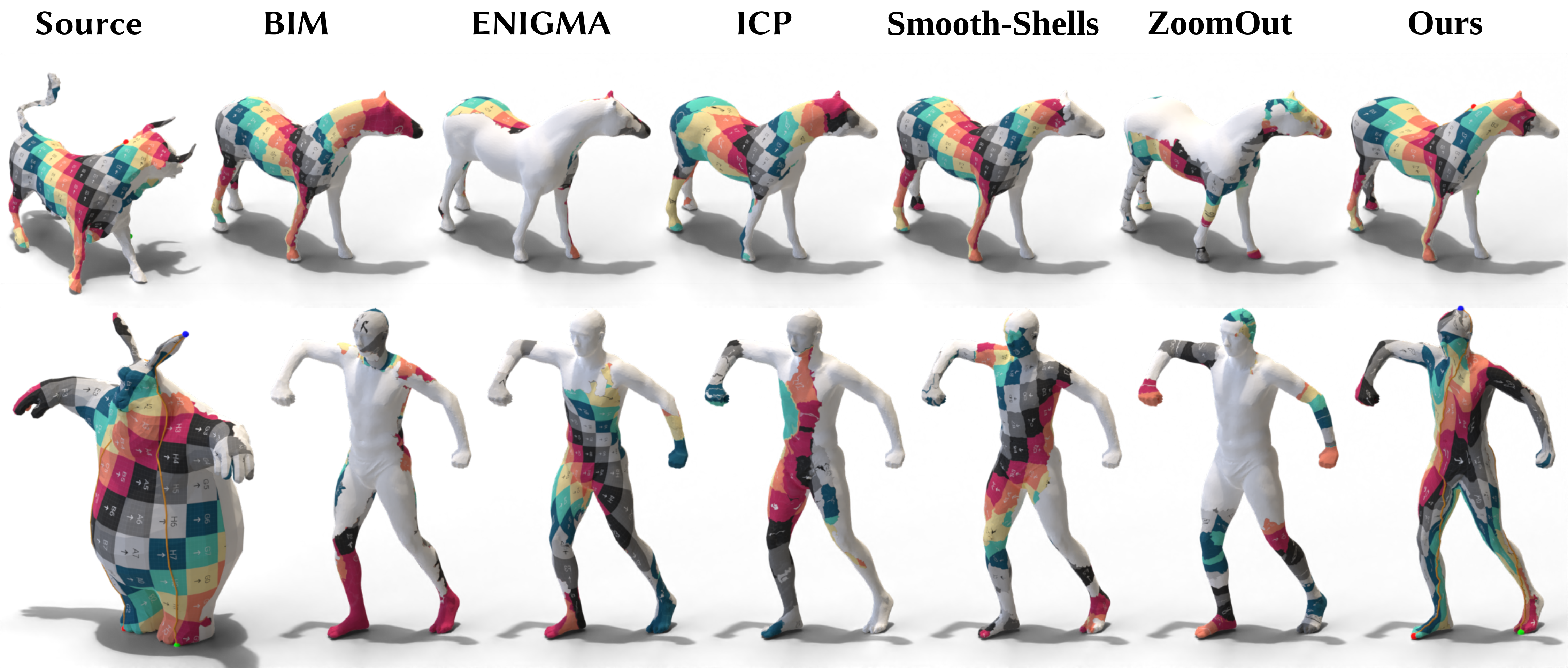}
        \caption{
    \textbf{Qualitative comparison.}\label{fig:enigma}
    ENIGMA~\shortcite{Edelstein20Enigma} fails to produce correct mappings, in cases of extreme deformations. Similarly, other state-of-the-art methods may lack bijectivity or correct correspondence. Ours can better handle these cases, see Table~\ref{tab:quantitative} for quantitative comparison.
    Colored landmarks and paths show \emph{automatically} selected cones and cuts by our method.
}
\end{figure*}

\section{Evaluation} \label{sec:experiments}
We evaluated our method on various datasets for inter-surface mapping and compared it against multiple baselines that focus on obtaining surface-to-surface maps.

\begin{table}[b!]
    \centering
    
    \caption{
        \textbf{Quantitative evaluation.} \label{tab:quantitative}
        We compute each map's accuracy (i.e., average geodesic error) and averaged them over 30 shape pairs for each dataset.
        }

\resizebox{0.48\textwidth}{!}{%
\setlength{\tabcolsep}{1pt}
    \begin{tabular}{r||c|c|c|c|c|c|c|c|c}
          & \multicolumn{3}{c|}{FAUST} & \multicolumn{3}{c|}{SHREC07} & \multicolumn{3}{c}{SHREC19} \\
          & Inv $\DA$ & Bij $\DA$ & Acc $\DA$ & Inv $\DA$ & Bij $\DA$ & Acc $\DA$ & Inv $\DA$ & Bij $\DA$ & Acc $\DA$ \\
          
         \hline
         ICP           & 0.06 & 0.17 & 0.25 & 0.09 & 0.65 & \textbf{0.23} & 0.07 & 0.75 & \change{0.15} \\
         BIM           & 0.09 & 0.03 & 0.04 & 0.49 & 0.48 & \textbf{0.23} & 0.05 & 0.82 & \change{0.04} \\
         Zoomout       & 0.33 & 0.23 & 0.15 & 0.25 & 0.65 & 0.54 & 0.29 & 0.76 & \change{0.32} \\
         Smooth-shells & 0.01 & \textbf{0.00} & \textbf{0.01} & 0.03 & 0.72 & 0.26 & 0.01 & 0.83 & \textbf{\change{0.01}} \\
         Ours         & \textbf{0.00} & \textbf{0.00} & 0.13 & \textbf{0.00} & \textbf{0.00} & \textbf{0.23} & \textbf{0.00} & \textbf{0.00} & \change{0.11} \\
    \end{tabular}
    }

\end{table}

\paragraph*{Datasets} %
We assess maps' quality on available benchmarks comprising isometric and non-isometric shape pairs. (i)~We randomly select 30 pairs from FAUST~\cite{bogo2014faust}, containing isometric deformations and pose variations of human shapes. (ii)~We choose 30 random same-category shape pairs from SHREC07 \cite{giorgi2007shrec}, containing non-isometric deformations across multiple categories of shapes. (iii)~We also extract 30 random shape pairs among the listed test set of SHREC19 from Dyke \etal~\cite{dyke2019shrec}, containing a mix of isometric and non-isometric deformations.

To ablate the effect of pose variation, we use FAUST~\cite{bogo2014faust}, SCAPE \cite{anguelov2004correlated}, and TOSCA \cite{bronstein2008numerical}. 
To ablate the effects of rendering settings and rotation, we use FAUST \cite{bogo2014faust}; 3DBiCar \cite{luo2023rabit}, which comprise a variety of textured shapes; and SHREC15 \cite{lian2015shrec}, which contain significant non-isometric-variations, with manually-annotated sparse correspondences. In the supplementary material, we present additional ablations highlighting the crucial role of initial alignment, the method's robustness to mesh holes and nois, and discuss Dinov2 features. To summarize, significant misalignment negatively impacts matching quality; feature similarity does not reflect their matching accuracy; finally, the method effectively maps meshes with holes, \eg, scans.

All meshes used in our experiment are watertight and genus zero, and range from 11K to 90K faces. The shape pairs include a mix of some isometric and mostly non-isometric cases. %

\paragraph*{Metrics} 
We assess map quality (see also \cite{ren2018continuous}) based on their accuracy, bijectivity, and inversion as: 
\begin{itemize}
    \item \textit{Accuracy (Acc $\DA$):} measures the ability of the algorithms to respect ground-truth correspondences. We measure it as the geodesic distance normalized, as defined in \cite{Kim11BIM}, for each landmark.
    \item \textit{Bijectivity (Bij $\DA$):} measures the geodesic distance of all vertices mapped forward, and then back to the source mesh wrt their original position. A zero value indicates perfect bijectivity.
    \item \textit{Inversion (Inv $\DA$):} measures how often the map flips the surface as the percentage of inverted triangles; we compute it as the agreement of the normal of the mapped triangles wrt the faces on which the triangle vertices are mapped.
\end{itemize}

\paragraph*{Baselines}
We compare with three other techniques that focus on extracting maps between given surfaces: (i)~BIM~\cite{Kim11BIM}, (ii)~Zoomout~\cite{melzi2019zoomout}, and (iii)~Smooth-shells~\cite{eisenberger2020smooth}. We also include (iv)~ICP, which uses the closest points as correspondence, as a strawman approach that performs well in case of negligible pose variation. Results are presented in Table \ref{tab:quantitative}, and selection of the pairs shown in Figure~\ref{fig:comparison}.

We cast ICP as nearest neighbor search after rigid alignment. Specifically, we use our pipeline first to align each shape pair and then compute nearest neighbor \changeb{correspondences} for each point on the source to the target mesh. This approach may perform well for shapes in similar poses with low isometric deformations.

Additionally, we compare qualitatively to Enigma~\cite{Edelstein20Enigma} that uses genetic algorithms along with a combinatorial search to find a set of good sparse correspondence, which are then interpolated to a dense low-distortion map. While this method produces smoother and more semantic maps than other baselines, it still suffers from large and uneven distortions, see Figure \ref{fig:enigma}.

\subsection{Qualitative Evaluation}
Figure~\ref{fig:resultsPage} shows Neural Semantic maps extracted using our fully automatic approach. The produced maps accurately match semantic features despite the fuzzy aggregated correspondences being erroneous and confused by symmetries (\eg, mapping incorrect limbs). Ours also work well across dissimilar shapes. These non-isometric cases require introducing significant local stretching to preserve semantic correspondence. The extracted maps still exhibit low isometric distortion, where possible, while adhering to semantically meaningful correspondences.
\change{Yet, artifacts may arise (see Armadillo's leg in Figure~\ref{fig:resultsPage}) when the smoothing energy is not sufficient to balance the noisiness of matches.} 
State-of-the-art methods, such as ENIGMA~\shortcite{Edelstein20Enigma} or Smooth-shells~\shortcite{eisenberger2020smooth}, suffer from self-symmetry ambiguities, \eg, see \texttt{bull}-\texttt{horse} in Figure~\ref{fig:enigma}.

\begin{figure}[t!]
\centering 
\includegraphics[width=\columnwidth]{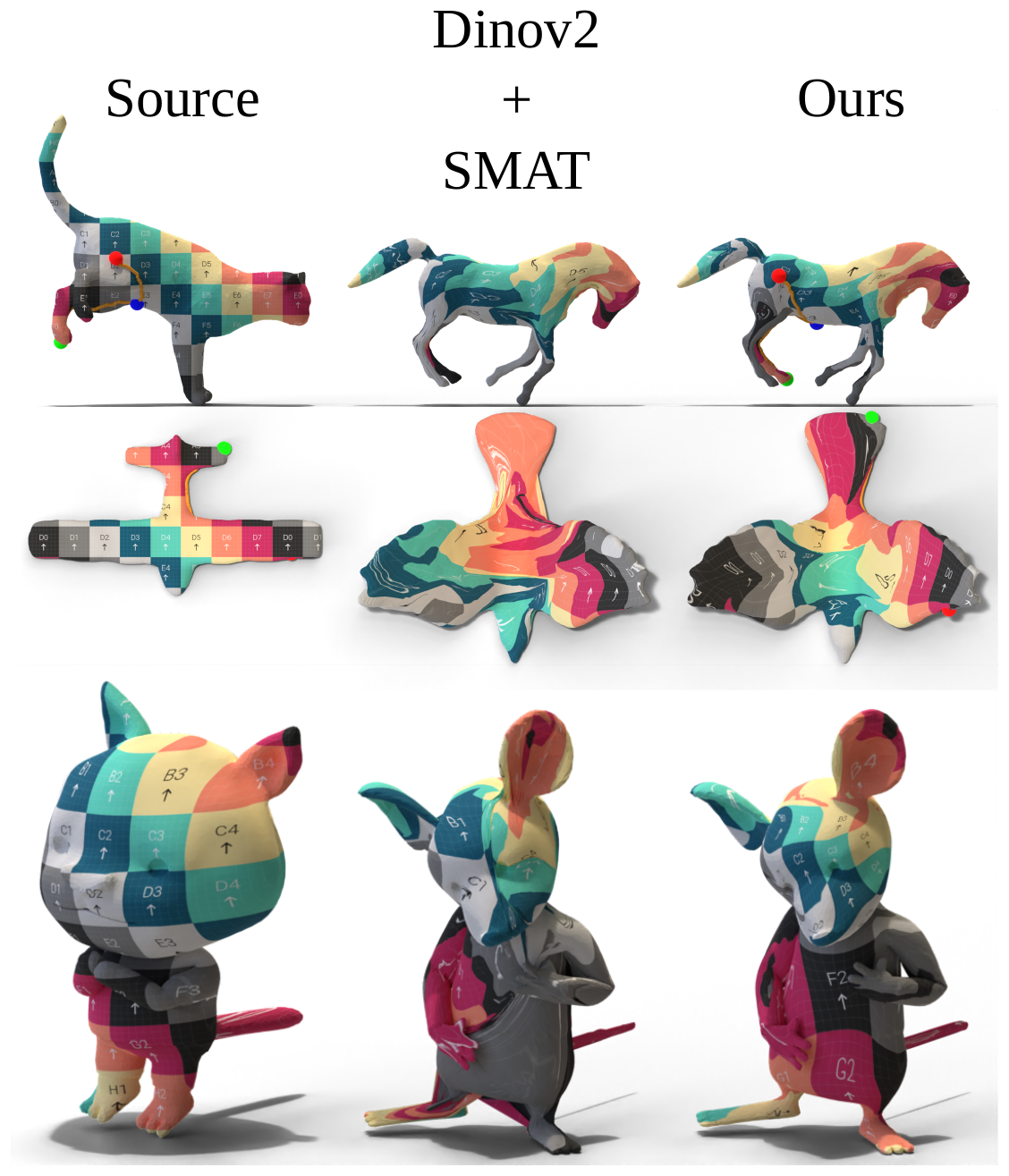}
        \caption{
    \textbf{Qualitative comparison.}\label{fig:smat}
    Surface Maps via Adaptive Triangulations (SMAT)~\shortcite{bijective:EG:23} optimize for bijective and continuous maps, relying on manual annotations. We pair it with Dinov2, Dinov2+SMAT, by replacing these manual annotations with $k=64$ automatically extracted ones, $\{\inmap^i\}$ with $i=1:k$, then we optimize the inter-surface map to construct an automatic inter-surface map. While Dinov2+SMAT attempts to satisfy all \changeb{correspondences} together with bijectivty, ours automatically filters out incorrect \changeb{correspondences}, yielding a more continuous and semantically correct map.
}
\end{figure}

\paragraph*{Aggregation}
To assess the importance of the map distillation module, we present a qualitative comparison in Figure \ref{fig:smat} with the method proposed by Surface Maps via Adaptive Triangulations (SMAT)~\cite{bijective:EG:23}, where we replace manual correspondences with automatically extracted ones. As the original approach requires a set of bijective \changeb{correspondences}, we randomly subsample a set of $N=64$ matches from the automatically extracted ones to ensure consistency, \ie, no vertex appears twice. Then, we optimize for a bijective map that respects these landmarks. We refer to it as Dinov2+SMAT. Note SMAT\shortcite{bijective:EG:23} optimize for isometric energy (Dirichlet), while we optimize only for smoothness, see Eq \ref{eq:symmetric_dirichlet}.

As SMAT does not account for inaccurate nor imprecise correspondences, it is unable to filter out wrong \changeb{correspondences}.  In our observations, optimizing a map with the original hyperparameters leads to visible inversions. This issue arises from SMAT's attempt to preserve all landmarks, resulting in maps with extreme stretches, a phenomenon intensified by the discrete nature of meshes. Adaptive remeshing struggles to handle these extreme stretches effectively, leading to visibly distorted maps. To mitigate this effect, we trade landmark precision for map continuity and quality. As shown in Figure \ref{fig:smat}, although both maps appear continuous, \cite{bijective:EG:23} is unable to filter out inaccurate \changeb{correspondences} and yield a reasonable map. Note that this experiment mainly assesses the importance of our correspondence distillation step, and Dinov2+SMAT is not the mode SMAT was originally designed for. Ours, without any explicit isometric or conformal energy term, still can produce smooth and semantics-respecting bijective maps. 

\subsection{Quantitative Evaluation}
We report quantitative errors using the metrics discussed earlier. 
In particular, for accuracy, we follow the standard practice and measure the mean geodesic distance to ground truth correspondence on a unit-area mesh. %

Although not guaranteed by construction, we empirically found that \change{ours consistently offers} more bijective and continuous maps, see Table \ref{tab:quantitative} "\textit{Bij}" and "\textit{Inv}", while others can fail to perfectly achieve these properties in both isometric and non-isometric cases. Our technique shows comparable quality in the maps \change{in non-isometric cases (SHREC07)} compared to state-of-the-art methods, Table \ref{tab:quantitative} "\textit{Acc}"\change{, while it performs worse in isometric cases (FAUST and SHREC19). In general, our method suffers in these cases} as it does not exploit geometric cues and does not have an explicit isometric energy term, thus producing less accurate maps than competing methods.

\section{Limitations}

\paragraph*{Timing.} 
A key limitation of our method is its long running time. The map optimization takes on average 1.5 hours, converting the meshes into their neural representation \change{which requires about 1 hour}, and extracting all Dino-ViT matches takes 21 minutes. We plan to investigate approaches such as Meta-Learning and better caching to speed up this process.

\paragraph*{Occlusion.}
The presence of self-occlusion in shape pairs prevents DinoViT from correctly mapping regions across shapes, thus consistently making mistakes. We believe incorporating other priors, or an advanced rendering pipeline (e.g., layered rendering) may help cope with this issue.

\paragraph*{Thin parts.}
We struggle to handle thin parts, as our pipeline requires parameterizing objects. Specifically, thin parts are difficult to handle unless \change{cut points} are manually placed.

\section{Conclusion}

We have presented a method that produces a semantic surface-to-surface map guided by visual semantic priors, by computing it from a set of candidate non-injective and discontinuous partial maps extracted by matchings over renderings of untextured 3D surfaces. Our method has many potential practical applications, ranging from matching scans of human faces and bodies to clothes, anatomical scans, and archaeological findings. These depend on the quality of the matchings achieved over the renderings of objects from these categories, which we aim to explore.

\paragraph*{Future work.}
We require surfaces to be cut as required by NSM~\cite{morreale2021neural} which makes our method more prone to error. We aim to improve the existing pipeline to avoid cutting altogether by replacing the 2D disks with 3D spheres~\cite{sphericalParam:03}, as successfully used in \cite{bijective:EG:23}. 
Our optimization cannot guarantee achieving a global optimum nor that the global optimum defines the ``most-meaningful'' semantic map, and we mark extending our method to directly \emph{learn} to produce maps from a dataset as an important future direction. Our method can create such a dataset, augmented with manual input to score the goodness of any extracted semantic map. 
We believe this work is only a step in producing semantic-driven maps. Candidate fuzzy maps extracted from other means can be considered. For instance, methods to predict fuzzy geometric correspondences directly over 3D surfaces trained for specific tasks can alternatively produce fuzzy maps and can be used in conjunction with semantic and/or visual cues.

\bibliographystyle{eg-alpha-doi} 
\bibliography{references}

\newpage
\appendix

\appendix

\section{Pseudocode}

We provide pseudocode for our semantic homeomorphic map extraction framework in 
Algorithm~\ref{alg:main}.

\RestyleAlgo{ruled}

\SetKwComment{Comment}{/* }{ */}

\begin{algorithm}[h!]
    \caption{Semantic Surface Homeomorphism}\label{alg:pipeline}
    \label{alg:main}
    \KwData{source $\source$, target $\target$}

    $ \mathbf{R} \gets \textsc{coAlign}(\text{DinoViT}(), \source, \target)$ \;
    $ \texttt{fuzzyMatches} \gets \textsc{computeMatches}(\text{DinoViT}(), \source, \target, \mathbf{R})$ \;
    $ A_{disk}, B_{disk} \gets \textsc{asyncCut}(\source, \target, \texttt{fuzzyMatches}) $ \;
    $ A_{NSM} \gets \textsc{overfitNSM}(A_{disk}) $ \;
    $ B_{NSM} \gets \textsc{overfitNSM}(B_{disk}) $ \;
    $\texttt{map} \gets \textsc{distilMap}(A_{NSM}, B_{NSM}, \texttt{fuzzyMatches})$ \;
    \text{return} $\texttt{map}$
    
\end{algorithm}

\section{Rendering Details}
In all cases, we render images of the same size, i.e., $1344 \times 1344$ with Mitsuba~\cite{Mitsuba3} using $spp=150$ and a path integrator.
When extracting semantic matches, we limit to rotations around the up-axis (y) - 20 steps between $[0, 2\pi)$ - and forward-axis (z) - 10 steps between $[\frac{-\pi}{2}, \frac{\pi}{2})$ - obtaining 200 images for each shape. Similarly, to align shapes, we rotate around the up-axis - 12 steps - with fixed increments.
To uplift 2D pixels to 3D for the matches, we use ray-triangle intersection. On average, we get $328$ \changeb{correspondences} per view, totaling $65$k \changeb{correspondences} across the $200$ views.

\section{Computing rendering correspondences}
As discussed in the main manuscript, we render the two surfaces from a given viewpoint to get two renderings, $R^\source_V$ and $R^\target_V$. We leverage DinoV2 \cite{oquab2023dinov2} to extract semantic features in the image space, thus obtaining $\lambda^\source_{i}$ and $\lambda^\target_{i}$ as features of rendering of $R^\source_V$ and $R^\target_V$, respectively. 
Then, to segment foreground/background we rely on PCA's first component of these features as it naturally groups them in opposite half-spaces. %

Finally, we match features with the cosine similarity between all feature pairs from the same viewpoint, as score $S_{ij}$. We define the match of patch $i\in R^\source_V$  as the patch $j\in R^\target_V$ with the highest cosine similarity, and vice versa, the match of patch $j\in R^\target_V$ as the patch $i\in R^\source_V$ with the highest cosine similarity. In summary, the pair $(i,j), i\in R^\source_V,j\in R^\target_V$ is a match, if 
\begin{equation}
S_{ij} = \max_k S_{ik} \text{ or } S_{ij} = \max_l S_{lj}.
\end{equation}

\change{
\subsection{Patch generation, feature extraction, and PCA}
Images are split into (non-overlapping) patches of 14 pixels. Then, Dinov2\cite{oquab2023dinov2} embeds these patches in a forward pass. Following \cite{amir2021deep}, we use \emph{keys} as feature vectors.
}

\change{
To segment foreground/background we rely on PCA's first component of the features. As discussed in \cite{oquab2023dinov2}, the features' sign naturally groups them in opposite half-spaces. As the sign is appointed randomly, we use the attention mask from the last layer to select the correct half-space: we average the first PCA component of the features and take the half-space which agrees with the positive attention mask. Matches are estimated only between foreground patches.
}

\change{
Finally, to unproject a match to 3D, we first translate a patch to a pixel using the known patch size, and then identify the 3D point on each shape (ray casting). }

\begin{figure}[!ht]
\centering
\begin{tikzpicture}[scale=0.9]

    \begin{axis}[%
            xlabel=Misalignment ($^\circ$),
            ylabel=AVG geodesic error,
        scatter/classes={%
        a={mark=o,draw=red},b={mark=x,draw=blue}}]
        \addplot+[
            scatter,only marks,%
            scatter src=explicit symbolic]%
            table[meta=label] {
                    x y label
                      0 0.11 a
                     20 0.12 a
                     40 0.15 a
                     60 0.18 a
                     80 0.22 a
                    100 0.26 a
                    120 0.29 a
                    140 0.33 a
                    160 0.35 a
                    180 0.35 a
                    200 0.35 a
                    220 0.33 a
                    240 0.30 a
                    260 0.26 a
                    280 0.23 a
                    300 0.19 a
                    320 0.15 a
                    340 0.12 a
            };
        \addlegendentry{L9 Dino V2}
        \addplot[
            scatter,only marks,%
            scatter src=explicit symbolic]%
            table[meta=label] {
                x y label
                  0 0.11 b
                 20 0.11 b
                 40 0.14 b
                 60 0.17 b
                 80 0.21 b
                100 0.25 b
                120 0.29 b
                140 0.29 b
                160 0.32 b
                180 0.34 b
                200 0.35 b
                220 0.34 b
                240 0.31 b
                260 0.27 b
                280 0.24 b
                300 0.20 b
                320 0.16 b
                340 0.13 b
            };
        \addlegendentry{L11 Dino V2}
    \end{axis}
\end{tikzpicture}

    \caption{
    \textbf{Robustness to misalignment:} \label{fig:ablation_rot}
    the quality of matches depends on the quality of alignment. In the case of severe misalignment ($60^\circ$ or more), we observe poor correspondence.
    }
\end{figure}
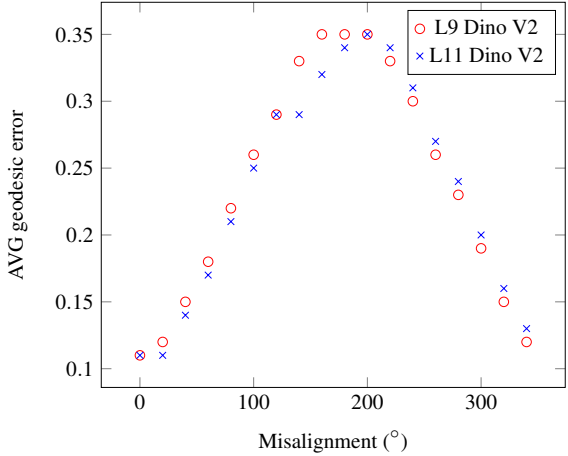

\begin{table}[b!]
    \centering
    \caption{
        \textbf{Dino ViT pose ablation:} \label{tab:vit_pose}
        DinoV2\shortcite{oquab2023dinov2} matches are significantly more accurate than DinoV1\shortcite{caron2021emerging} in case of pose variation, with no significant difference between features from L9 and L11.
    }
    \begin{tabular}{r|c|c|c|c|c|c}
                   & \multicolumn{2}{c|}{\coloredtext{Dense}{FAUST}} & \multicolumn{2}{c|}{\coloredtext{Dense}{SCAPE}} & \multicolumn{2}{c}{\coloredtext{Dense}{TOSCA}} \\
        Layer      & 9 & 11 & 9 & 11 & 9 & 11 \\
        \hline
        DinoV1     & 0.16 & 0.16 & 0.38 & 0.40 & 0.27 & 0.29 \\
        DinoV2     & 0.09 & 0.09 & 0.18 & 0.18 & 0.27 & 0.25 \\
        
    \end{tabular}
    
\end{table}

\begin{figure}[t]
\centering 
    \includegraphics[width=\columnwidth]{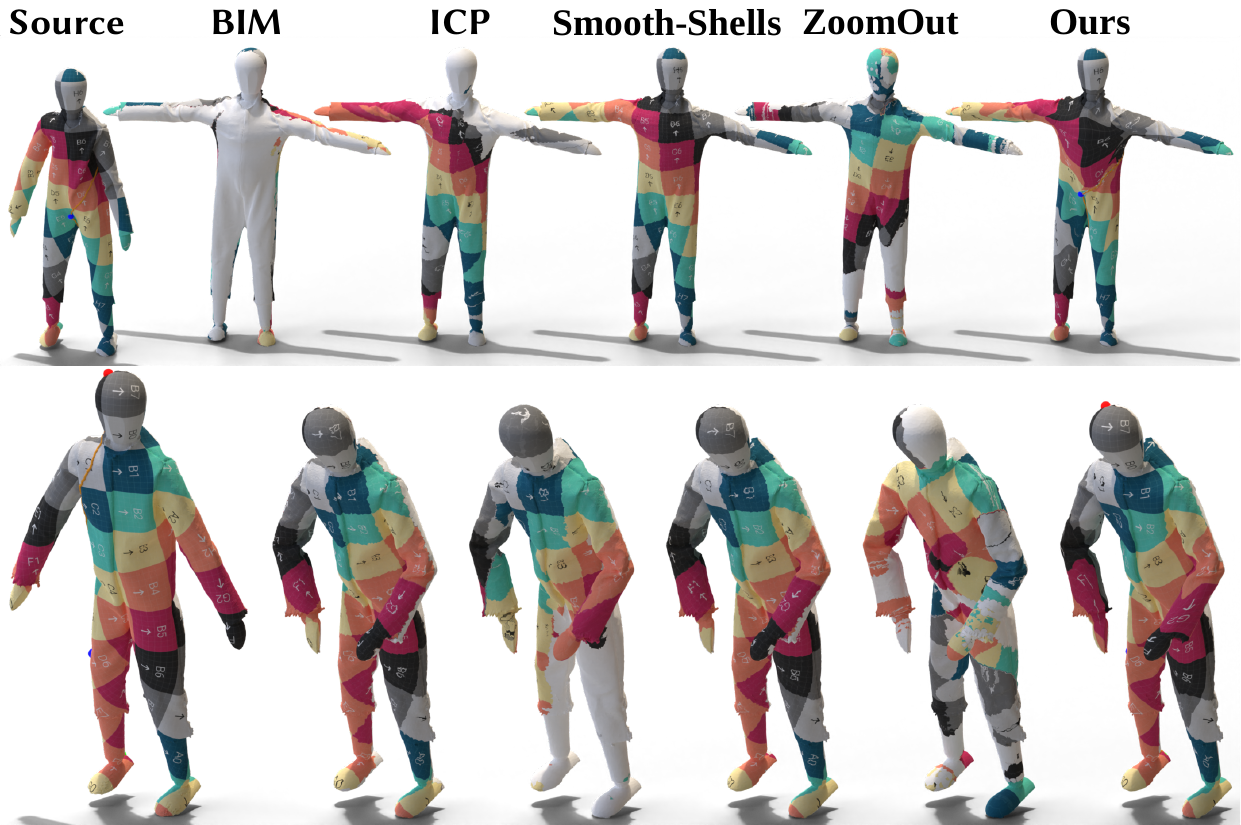}
        \caption{
    \change{
    \textbf{Qualitative comparison SHREC19:}\label{fig:shrec19}
    Functional maps-based methods produce good maps, although often being discontinuous. Ours explicitly encourages continuity and bijectivity.
    }
}
\end{figure}

\section{Comparison Details}
We discuss the main considerations for/against the competing algorithms we compare against.

Blended Intrinsic Maps (BIM)~\cite{Kim11BIM} is a classic method that uses geometric priors without any learning component. Namely, it picks a subset of self-consistent and low-distortion conformal maps and then blends them using weighted averages. Individual conformal maps can handle very non-isometric surfaces, however, they can produce high isometric distortion even in near-isometric cases. Note also that the resulting blended map is not a homeomorphism nor even continuous.

Zoomout~\cite{melzi2019zoomout} and Smooth-shells~\cite{eisenberger2020smooth} are both functional maps-based methods. Zoomout starts with a small functional correspondence matrix and iterative upsamples it in the spectral domain. Smooth-shells follow a similar coarse-to-fine scheme, relying on shells as a proxy for functional basis. To handle self-symmetries, Eisenberger et al.~\shortcite{eisenberger2020smooth} incorporate MCMC to evaluate multiple possible functional maps. 

\change{We initialize Zoomout's map ($C_{21}$) as an identity of size $4$ as by official implementation. Then, we refine it until it contains $50$ eigenvectors.}
\change{Similarly, for Smooth-shells we follow the official implementation and use MCMC to bootstrap the map using $K_{min}=6$ and $K_{max}=20$. and evaluating $N_{prop}=500$ proposal. In both cases, no landmarks are used.
Finally, for ICP we first align the two input shapes as described in Sec. 3.1, and then estimate the nearest neighbor for each vertex.}

\change{
We depict maps for the different methods on SHREC19 in Figure \ref{fig:shrec19}. State-of-the-art methods work well as they exploit geometric cues, although they are susceptible to self-symmetries (see BIM\cite{Kim11BIM} first row). Conversely, "Ours" relies purely on visual cues, with no isometric regularization, thus being less accurate on average.
}

\change{
\section{Differences with Neural Surface Maps}
Neural Surface Maps~\cite{morreale2021neural} defines the general mapping framework used to optimize maps. Following the original work, the input two shapes must be homomorphic to a disk with their boundary in correspondence. As this constraint is impossible to satisfy automatically, this work relies on seamless maps, thus relaxing this constraint to $3$ corresponding points which are extracted automatically. Furthermore, we define a soft correspondence term to handle inaccurate correspondences, while NSM enforces exact correspondences with an L2 loss over all correspondences.
}

\change{
\section{Metrics}
In all experiments, all shapes are automatically normalized and centered.
}

\change{
\paragraph{Bijectivity}
We estimate the map's bijectivity of the shape vertices for all baselines. For ICP, BIM, Zoomout, and Smooth-shells we map all vertices forward ($A \rightarrow B$) and then backward ($A \leftarrow B$), using the forward and backward map respectively. Then, we compute the geodesic distance between the starting vertex and its forward-backward map.}

\change{Similarly, for consistency we evaluate "Ours" bijectivity only for the shape vertices. In particular, we map a vertex in A onto B’s 2D domain through $h$, and then, we use the piecewise linear map for 2D-3D. For B to A, we pullback vertices through barycentric coordinates after mapping forward all A’s triangles. Empirically, for "Ours" \emph{we never observe flips}; while for baselines, correspondences are always given, thus, no ambiguity arises. In the case of a non-bijective map, we would consider the first triangle.
}

\begin{table}[t]
    \centering

    \caption{
        \textbf{Dino ViT ablation:} \label{tab:vit_ablation}
       DinoV2\shortcite{oquab2023dinov2} works better than its predecessor\shortcite{caron2021emerging}, with no significant difference between features from L9 and L11. The use of colored lights (rows DinoV1 and DinoV2) offers better visual cues to extract matches than white lights. Although counter-intuitive, the use of \emph{simple} texture reduces the visual cues available to Dino ViT.
    }
    \resizebox{\columnwidth}{!}{
    \begin{tabular}{r|c|c|c|c|c|c}
                   & \multicolumn{2}{c|}{\coloredtext{Dense}{FAUST}} & \multicolumn{2}{c|}{\coloredtext{Sparse}{SHREC15}} & \multicolumn{2}{c}{\coloredtext{Sparse}{3DBiCar}} \\
        Layer      & 9 & 11 & 9 & 11 & 9 & 11 \\
        \hline
        DinoV1     & 0.10 & 0.12 & 0.32 & 0.32 & 0.36 & 0.49 \\
        \hline
        DinoV2     & 0.11 & 0.11  & 0.24 & 0.24 & 0.33 & 0.33 \\ 
        
        \hline
        white lights (V1) & 0.20 & 0.18 & 0.27 & 0.35 & 0.38 & 0.38 \\ 
        \hline
        white lights (V2) & 0.11 & 0.11 & 0.24 & 0.24 & 0.30 & 0.31 \\ 
        \hline
        texture (V1) & - & - & - & - & 0.26 & 0.26 \\
        \hline
        texture (V2) & - & - & - & - & 0.29 & 0.29 \\
        
    \end{tabular}
    }

\end{table}

\begin{figure}[b!]
\centering
    \includegraphics[width=\columnwidth]{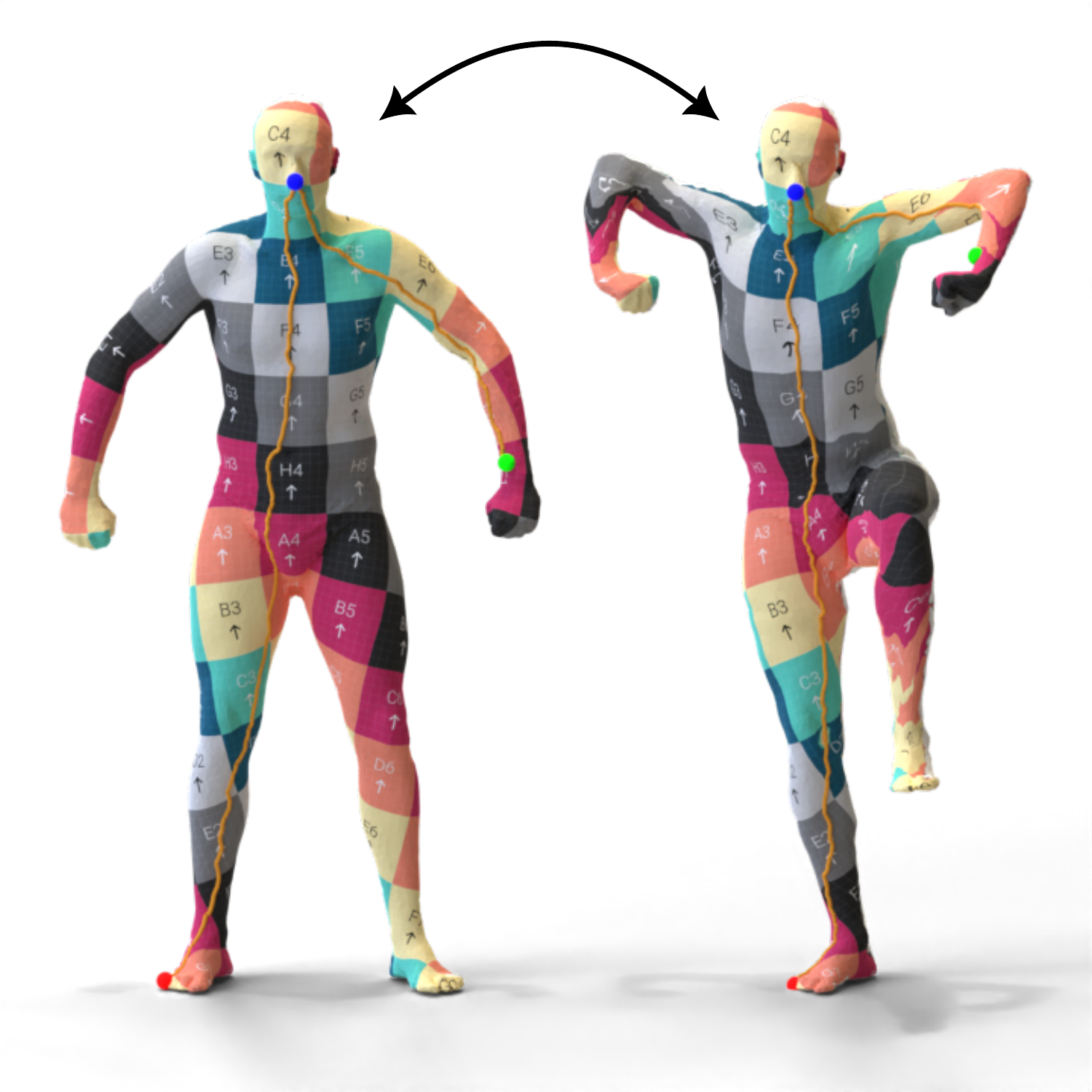}
    \caption{
    \textbf{Pose variation:}\label{fig:pose}
    we assess the ability of DinoV2~\shortcite{oquab2023dinov2} to establish matches between shapes in different poses, as those in the figure. Experimentally, DinoV2 yields correspondences able to guide our pipeline to a proper solution. Colored landmarks and paths show \emph{automatically} selected cones and cuts by our method.
}
\end{figure}

\begin{figure}[t]
\centering
         \includegraphics[width=\columnwidth]{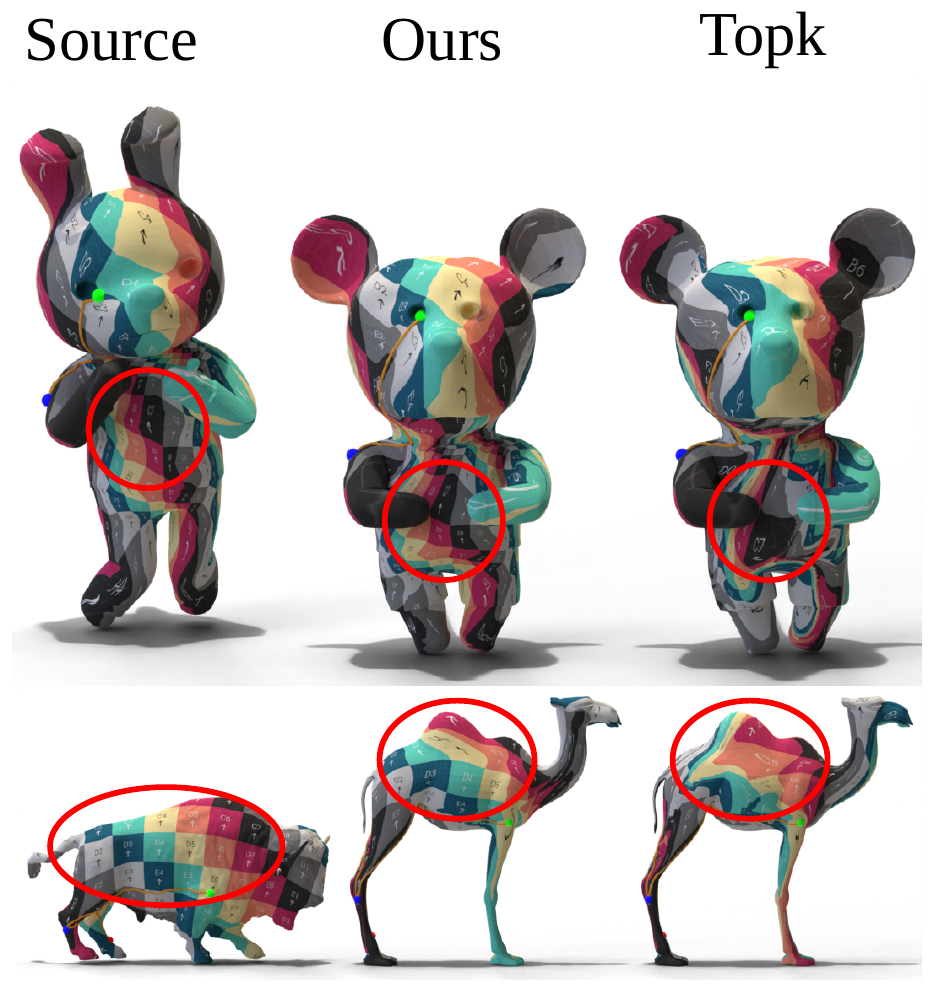}
    \caption{
    \textbf{Similarity scores.}\label{fig:similarity}
    \textbf{Right}: a map optimized with the top $k=100$ correspondences based on the similarity score. \textbf{Middle}: map optimized using all correspondences. \textbf{Left}: source mesh.
    The map optimized with matches with the highest similarity score matches shows several incorrectness, highlighted with a red circle. This is the result of several incorrect matches which bias the map towards an incorrect energy minimum. Differently, using all correspondences prevents this behavior, as the optimization process automatically filters out wrong matches.
}
   \vspace{-0.15in}
\end{figure}

\section{Ablation}

\subsection{On Dinov2 features}
As aforementioned, we deem a match if the cosine similarity $S_{ij}$ between patch features - $\lambda^\source_{i}$ and $\lambda^\target_{i}$ - is the highest. While this is a common similarity measure, it is important to acknowledge its inherent limitations. Specifically, one notable challenge is that similarity scores derived from different images may not be directly comparable. For example, for two \changeb{correspondences} with scores $0.9$ and $0.8$, the former match pair is not necessarily better than the latter. In essence, features extracted from one view may be extremely dissimilar to those extracted from another view, even for the same shape. This arises from the inherent variation in image structure across different views and how features are generated from them. 
This inherent variability hinders consistency in cross-image feature comparisons. Consequently, the process of aggregating features across different views can potentially yield unexpected outcomes, leading to either incorrect matching or highly inaccurate results. 

Experimentally, sampling the top $k=100$ correspondences based on the similarity produced far worse results than uniform sampling or uniform weighting, see Figure \ref{fig:similarity} for qualitative comparison. In both cases, we optimize maps following the proposed algorithm: \emph{Ours} uses all \changeb{correspondences}, while \emph{TopK} is limited to $k=100$ correspondences with the highest similarity score. Visibly, some of these correspondences are incorrect and bias the map towards incorrect minima, thus their similarity score is not representative of their quality. Indeed, the use of all \changeb{correspondences} prevents the map from falling into such a degenerate solution, as the majority of \changeb{correspondences} are reasonably correct.

\subsection{Tuning DinoVit Matches}
We ablate the quality of matches based on DinoViT's degrees of freedom - layer features - in different contexts: pose variation, presence of texture, lights, and misalignment. We conduct our analysis on three distinct datasets: \coloredtext{Dense}{FAUST}~\cite{bogo2014faust}, \coloredtext{Sparse}{3DBiCar}~\cite{luo2023rabit}, and \coloredtext{Sparse}{SHREC15}~\cite{lian2015shrec} each with \coloredtext{Dense}{dense} or \coloredtext{Sparse}{sparse} ground truth.

We select 12 shape pairs, 4 for each dataset, to ablate texture and misalign concerning the choice of Dino ViT feature layer, as discussed in \cite{amir2021deep}. Similarly, we assess the effect of pose variation for the same model with a single instance of FAUST, SCAPE, and TOSCA mapped onto all the other provided poses. We report the quantitative results in Table \ref{tab:vit_ablation} and show shape pairs examples and qualitative optimization results in Figure \ref{fig:pose}.

We assess the quality of the aggregated \changeb{correspondences} in terms of the normalized average geodesic distance~\cite{Kim11BIM}. We follow the procedure described in the main paper to aggregate the fuzzy \changeb{correspondences}, thus, obtaining a face-wise map $M$ from one mesh onto the other. Finally, the geodesic distance is computed on the target mesh between the centroid of the mapped face to the centroid of the ground truth target face.

In general, DinoV2~\cite{oquab2023dinov2} outperforms its predecessor V1~\shortcite{caron2021emerging}, offering more accurate and robust matches. The depth at which features are extracted (9 vs 11) does not impact the matches of DinoV2, while it plays a significant role for DinoV1, as discussed in \cite{amir2021deep}. 
The presence of texture is beneficial to DinoV1, while it only offers a minor improvement for DinoV2. This is reassuring as our method can only assume access to untextured models. 
The choice of colored lights offers additional shading and visual features for DinoV1, but it is less relevant for DinoV2 as white lights perform equally with the base case.

\subsection{Effect of Initial Alignment}
We ablate the effect and robustness to misalignment for \changeb{correspondences} quality, see Figure \ref{fig:ablation_rot}.
We start from a correct alignment with 12 shape pairs and incrementally misalign one shape - step of 20$^\circ$ around the up axis. We report the quality of \changeb{correspondences} in terms of geodesic error, i.e., accuracy.
The quality sensibly decreases with severe misalignment - more than 40$^\circ$ - reaching a peak with opposite orientation - 180$^\circ$.
We additionally compare the quality of \changeb{correspondences} for the last two layers of Dino-ViT and show that, for such a case, a deeper level (L11) seems to encode slightly better semantic information than the previous layer (L9).

\subsection{Handling Noise and Holes}
\begin{figure}[t]
\centering
    \includegraphics[width=\columnwidth]{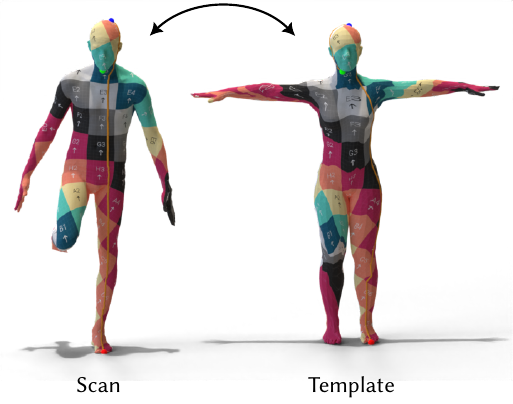}
    \caption{
    \textbf{Scan to SMPL:}\label{fig:scan}
    we first close holes in the raw scan (left) with Meshlab\shortcite{meshlab}, then we map it onto the template SMPL model\shortcite{loper2015smpl} and mask out the surfaced introduced to fill holes. Colored landmarks and paths show \emph{automatically} selected cones and cuts by our method.
}
\end{figure}

Raw scans present noise or holes, thus inhibiting the applicability of our method since it assumes watertight genus zero meshes. Intuitively the presence of large holes, and missing limbs such as arms, may severely mislead DinoViT and thus our pipeline. On the other hand, small holes can be dealt with by applying a simple hole-filling approach. In Figure \ref{fig:scan}, we use our method to map a raw scan to the SMPL template~\cite{loper2015smpl}. We prefill small holes with Meshlab~\cite{meshlab} and then apply our pipeline.

\end{document}